\documentclass[10pt,twocolumn,letterpaper]{article}
\usepackage[pagenumbers]{cvpr}
\usepackage{graphicx}
\usepackage{amsmath}
\usepackage{amssymb}
\usepackage{booktabs}

\usepackage{float}
\usepackage{algorithm}
\usepackage{algorithmic}
\usepackage[pagebackref,breaklinks,colorlinks,citecolor=cvprblue]{hyperref}
\definecolor{cvprblue}{rgb}{0.21,0.49,0.74}

\title{DIG-FACE: De-biased Learning for Generalized Facial Expression Category Discovery}
\newtheorem{definition}{Definition}

\newtheorem{Proof}{Proof}
\newtheorem{lemma}{Lemma}

\begin{document}

\author{%
  \fontsize{10pt}{12pt}\selectfont
  Tingzhang Luo\textsuperscript{1}\footnotemark[2] \quad
  Yichao Liu\textsuperscript{1}\footnotemark[2] \quad
  Yuanyuan Liu\textsuperscript{1}\footnotemark[1] \quad
  Andi Zhang\textsuperscript{2} \\  \fontsize{10pt}{12pt}\selectfont
  Xin Wang\textsuperscript{3} \quad
  Yibing Zhan\textsuperscript{4} \quad
  Chang Tang\textsuperscript{1} \quad
  Leyuan Liu\textsuperscript{5} \quad
  Zhe Chen\textsuperscript{6} \\  \fontsize{10pt}{12pt}\selectfont
  \textsuperscript{1}China University of Geosciences \quad
  \textsuperscript{2}University of Cambridge \quad
  \textsuperscript{3}Baidu Inc \\  \fontsize{10pt}{12pt}\selectfont
  \textsuperscript{4}JD Explore Academy \quad
  \textsuperscript{5}Central China Normal University \quad
  \textsuperscript{6}La Trobe University \\
  {\fontsize{9pt}{11pt}\selectfont\url{https://github.com/Clarence-CV/G-FACE}}
}

\maketitle

\renewcommand{\thefootnote}{\fnsymbol{footnote}}
\footnotetext[1]{Corresponding author.}
\footnotetext[2]{Equal contribution.}


\begin{abstract} 
We introduce a novel task, \textbf{G}eneralized \textbf{FA}cial expression \textbf{C}ategory discov\textbf{E}ry (G-FACE), 
that discovers new, unseen facial expressions while recognizing known categories effectively. 
Even though there are generalized category discovery methods for natural images, they show compromised performance on G-FACE. We identified two biases that affect the learning: 
\textbf{implicit bias}, coming from an underlying distributional gap between new categories in unlabeled data and known categories in labeled data, and \textbf{explicit bias}, coming from shifted preference on explicit visual facial change characteristics from known expressions to unknown expressions. 
By addressing the challenges caused by both biases, we propose a Debiased G-FACE method, namely \textbf{DIG-FACE}, that facilitates the de-biasing of both implicit and explicit biases. 
In the implicit debiasing process of DIG-FACE, we devise a novel learning strategy that aims at estimating and minimizing the upper bound of implicit bias. 
In the explicit debiasing process, we optimize the model’s ability to handle nuanced visual facial expression data by introducing a hierarchical  category-discrimination refinement strategy: {sample-level}, {triplet-level}, and {distribution-level} optimizations. 
Extensive experiments demonstrate that our DIG-FACE significantly enhances recognition accuracy for both known and new categories, setting a first-of-its-kind standard for the task. 



\end{abstract}

\section{Introduction}
\label{sec:intro}
Facial expression recognition (FER) is important in various tasks like the human-computer interaction, as expressions are a significant manifestation of human emotions~\cite*{li2020deep,kumari2015facial}. Traditional models are trained to recognize seven basic facial expressions, \textit{i.e., }happy, sad, surprise, anger, fear, disgust, and neutral. 
However, researches~\cite{Kollias_2023_CVPR,cowen2021sixteen,liu2022mafw} indicate that humans can exhibit expressions that extend beyond these basic categories, including composite and complex expressions, such as happy-surprise and perplexity. Sometimes, human expressions contain many complex and unexplored subtleties and variations~\cite{zhao2016facial}. 
When encountering new types of expressions in real-world scenarios, models trained solely on basic expressions would perform poorly. 
Furthermore, manually annotating each expression in open-world FER scenarios could be prohibitively expensive and impractical. Therefore, reducing manual annotations and discovering generalized facial expression categories have become increasingly necessary when exploring the open-set setting.

\begin{figure}[t]
    \centering
\includegraphics[width=0.88\linewidth]
{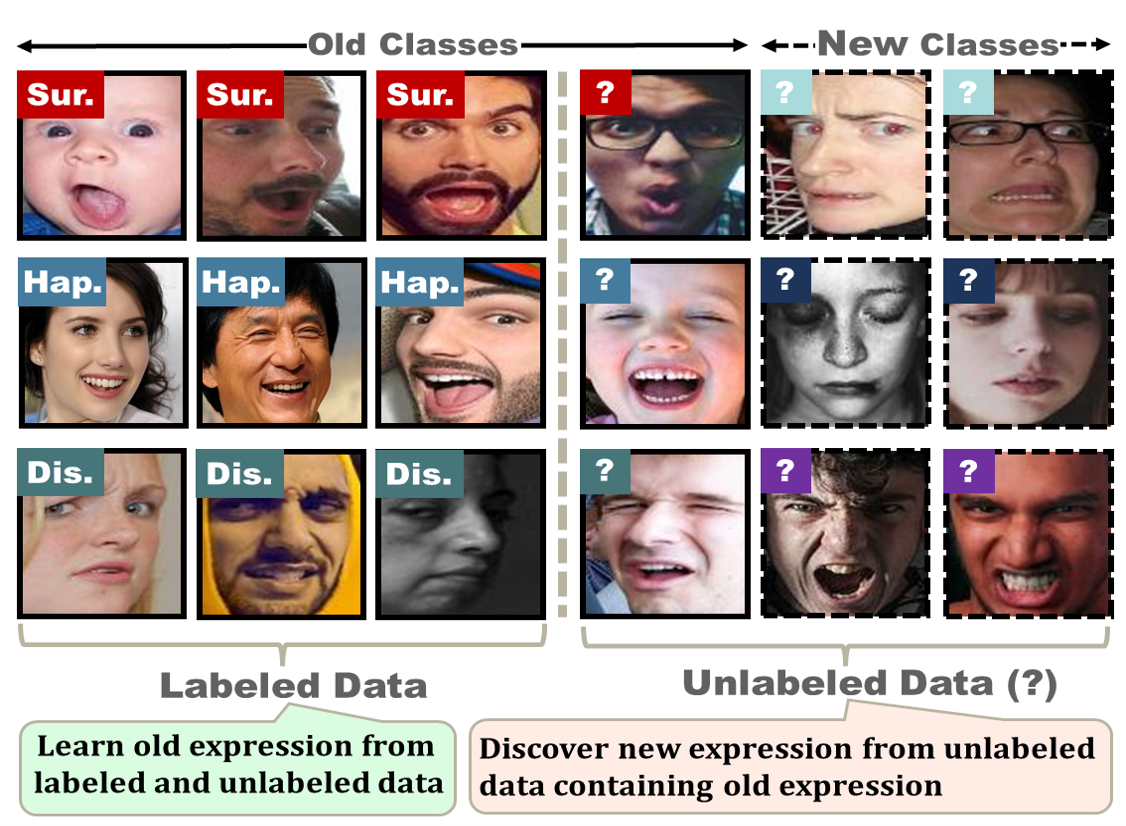}
    \captionof{figure}{
     G-FACE aims to discover unknown (new) facial expressions while recognizing known (old) classes, both of which are present in the unlabeled data.
    } \label{GCD}
\end{figure}

Recent advancements in open-set FER~\cite{zhang2024open,liu2024open} have improved the ability of models to detect new, unknown expressions while maintaining high accuracy on known categories. However, these methods primarily focus on identifying one unknown category alongside known categories, limiting their effectiveness in scenarios with multiple unknown expressions and hindering their applicability in real-world contexts.
Then, the need for addressing this limitation has driven the development of Generalized Category Discovery (GCD)~\cite{vaze2022generalized}. 
The GCD employs semi-supervised training to utilize both labeled and unlabeled data, assigning cluster IDs to the unlabeled data to enable recognition of both known and multiple unknown categories. Although GCD has shown success on natural image datasets~\cite{wang2024sptnet,luo2024contextuality}, we found that directly applying it to facial expression recognition (FER) introduces unique challenges. Unlike general images, FER data involves much more imbalanced distributions of known and unknown categories, as well as much more subtle variations between expressions, making it extremely difficult for current GCD methods to maintain accuracy on known categories while identifying new ones.
In this study, to specifically formulate the GCD task for FER, we draw inspiration from the typical GCD formulation and introduce a novel task, namely \textbf{G}eneralized 
\textbf{FA}cial expression \textbf{C}ategory discov\textbf{E}ry (G-FACE), to represent the task that simultaneous identifies known facial expressions and discovers new, unseen facial expressions in open-set data. 

\begin{figure}[t]
    \centering    \includegraphics[width=1\linewidth]{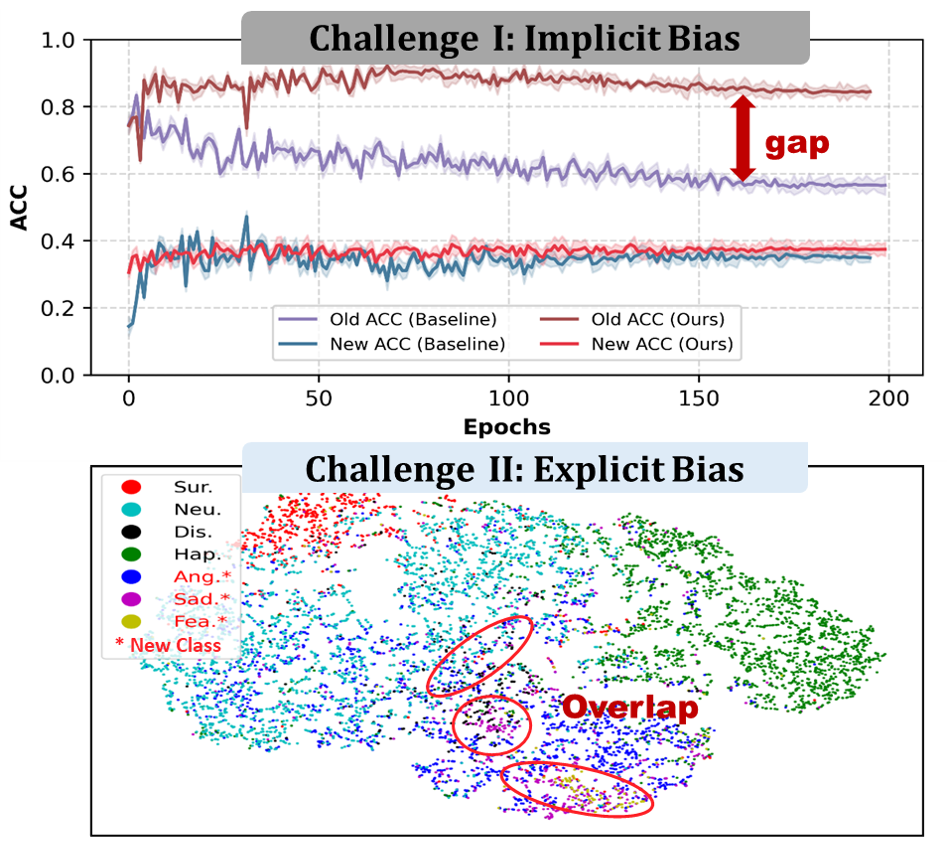}
    \caption{\textbf{Top}: 
Implicit bias leads to a decline in recognition accuracy for known categories in later training stages, as seen in previous GCD models like Baseline SimGCD.
\textbf{Bottom:} Explicit bias can be analyzed by observing the overlap between known and unknown classes, which may lead to a blurred decision boundary. We present a t-SNE visualization of the baseline SimGCD.}
    \label{fig:challenge}
\end{figure}

As mentioned above, the G-FACE presents unique challenges, which we could conclude as a biasing problem. More specifically, we observe implicit and explicit biases that would set obstacles to the G-FACE:  

\noindent$\bullet$ \textit{\textbf{Challenge I: Implicit Bias.}} During the semi-supervised learning for GCD, unlabeled data has new categories whose effective decision boundaries usually subtly shifted from known categories. This introduces implicit distributional gaps between known and unknown data, which increases the learning difficulties for the G-FACE. Fig. \ref{fig:challenge} (top) partially illustrates the effect caused by the implicit bias, where this shift leads to a significant drop in accuracy for known expression categories as training goes on (\emph{e.g.} possibly the effective decision boundaries of known have been shifted).


\noindent$\bullet$ \textit{\textbf{Challenge II: Explicit Bias.}} We identify explicit bias as the shifts in the preference of visible subtle facial characteristics from known expressions to unknown expressions. Different from implicit bias representing shifted decision boundaries, the differences in key facial characteristics between known and unknown expressions would make learned decision boundaries unclear and less distinguishable for both known and unknown categories. This can be partially supported by overlapped known/unknown data distributions, as shown in Fig. \ref{fig:challenge} (bottom) .

To address these issues, we propose a novel \textbf{D}e-b\textbf{I}ased \textbf{G-FACE} framework, \textbf{DIG-FACE}, which incorporates both implicit and explicit \emph{debiasing} strategies. In the implicit debiasing step, we introduce a novel \textit{F-discrepancy}-based metric to help reveal and reduce the upper bound of implicit bias. The explicit debiasing step enhances visual feature learning through a three-level category-discrimination refinement learning strategy to address fine-grained subtle distinctions. Together, these strategies significantly improve the model’s robustness and accuracy in the proposed G-FACE task. Our contributions can be summarized as:

\textcolor{orange}{(i)} We propose a novel G-FACE task, which aims to simultaneously recognize known facial expressions and discover unknown categories. To the best of our knowledge, this is the first time this problem has been formulated in the FER literature.

\textcolor{orange}{(ii)} We identify two core biases that introduce challenges in G-FACE: implicit bias from underlying distributional gap, and explicit bias from diversified preference on subtle visual facial characteristics, which may confuse decision boundaries of known/unknown categories, complicating accurate recognition.

\textcolor{orange}{(iii)} We propose DIG-FACE, incorporating both implicit and explicit debiasing processes to address the identified biases. The DIG-FACE has shown outstanding benefits on the G-FACE task, consistently outperforming state-of-the-art GCD approaches on various datasets by a great margin. 


\section{Related Work}
\noindent\textbf{Facial Expression Recognition (FER)}  conveys rich information and is a key focus of current AI research~\cite{ekman1993facial,tian2011facial,ruan2021feature,calder2016understanding,kanade2000comprehensive,li2020deep}. There have also been many studies in recent years aimed at improving the performance of FER models~\cite*{Yang_2018_CVPR,wang2020suppressing,li2022towards,zhang2024leave}. For instance, Yang \textit{et al}.~\cite{Yang_2018_CVPR} introduce a de-expression learning method. Wang and his team propose an effective Self-Cure Network (SCN). Besides, semi-supervised learning~\cite{li2022towards} and challenging imbalance problems~\cite{zhang2024leave} are also important research directions in FER. Recently, in order to further explore open-world settings, Zhang \textit{et al}.~\cite{zhang2024open} introduce open-set setting to FER scenario. In addition,  Liu \textit{et al}. proposed video-baset open-set setting in FER~\cite{liu2024open}. However the open-set task aims to detect new classes that do not belong to the previous known category without further categorization, which remains a limitation.\\
\noindent\textbf{Category Discovery} aims to discover new categories from unlabeled data. During training, labeled and unlabeled training data provide important visual conceptual information. 
The assumption of Novel Category Discovery (NCD)~\cite{fini2021unified,han2021autonovel,han2019learning,zhao2021novel} is that there is no overlap in categories between the labeled and unlabeled datasets. However, this assumption is unrealistic in the broader open-world~\cite{vaze2022generalized,pu2023dynamic,zhang2023promptcal} setting, where the unlabeled dataset not only encompasses categories previously learned by the model from labeled data but also novel categories. Prior studies have devised efficient techniques for category discovery, utilizing parametric classifiers~\cite{wen2023parametric,luo2024contextuality}, enhancing representation learning~\cite{pu2023dynamic,fei2022xcon}, or employing prompting learning with larger models~\cite{zhang2023promptcal}. 
In addition, some researches aim to focus on more realistic or difficult GCD problems, such as incremental learning-GCD~\cite{zhao2023incremental}, long-tail recognition GCD~\cite{li2023generalized}, cross-domain GCD~\cite{wang2024hilo}, \textit{etc}.
However, most existing GCD methods utilize a single shared classification head, thus failing to decouple the accumulation of biases.

\section{Problem Formulation}\label{task}
In this section, we formalize the task definition of G-FACE and provide a detailed explanation of its two major challenges: \textbf{Implicit Bias} and \textbf{Explicit Bias}. Tab. \ref{tab:notation} provides the preliminary notations.
\subsection{Problem Setting}
Following GCD setting \cite{vaze2022generalized}, the G-FACE problem focuses on preserving the model's capability to categorize known expression classes and simultaneously discover new ones from partially labeled and abundant unlabeled data. We define the unlabeled FER dataset as $\mathcal{D}_U=\{(\mathbf{x}_i^u, \mathbf{y}_i^u)\} \subset \mathcal{X} \times \mathcal{Y}_u$, where $\mathcal{Y}_u$ is the label space for unlabeled data points. The goal of G-FACE is to train a model that effectively categorizes the instances in $\mathcal{D}_U$ using information from a labeled FER dataset $\mathcal{D}_L=\{(\mathbf{x}_i^l, \mathbf{y}_i^l)\} \subset \mathcal{X} \times \mathcal{Y}_l$, where $\mathcal{Y}_l \subset \mathcal{Y}_u$. 
In this study, we adopt a parametric approach~\cite{wen2023parametric,luo2024contextuality,banerjee2024amend} and assume the number of categories in the unlabeled space is known, denoted as $K_u = |\mathcal{Y}_u|$. The evaluation metric is defined in \ref{metric}.

\subsection{Implicit Bias}
We identify the implicit bias which subtly shifts decision boundaries and creates distributional gaps. While this bias cannot be directly observed from the image data, it can be constrained through mathematical techniques. To address it, we define a discrepancy metric to measure the divergence between predictions and true labels and decompose the overall discrepancy to isolate the contribution of new categories. We then introduce the \textbf{\textit{F-discrepancy}}, a theoretical tool that characterizes the relationship between discrepancies on labeled and unlabeled data. Based on this, we derive a formal upper bound for implicit bias. The key notions are summarized in Tab \ref{tab:notation}.

\begin{definition}[Discrepancy Metric]\label{Metric}
To quantify the discrepancy between model predictions and true labels, we define a discrepancy metric $\xi(\cdot, \cdot)$. For a set of data points $\{x_i\}_{i=1}^{n}$, the model predictions are obtained through a softmax function and then mapped to Euclidean space $\mathbb{E}$ using a function $f: \mathbb{P} \rightarrow \mathbb{E}$:
\begin{equation}
\footnotesize
    \xi(\mathcal{H}_{h,\psi}(x),\mathcal{F}(x)) = 
    \sqrt{\sum_{i=1}^{n} \left\| \mathcal{H}(x_i) - \mathcal{F}(x_i) \right\|^2},
\end{equation}
where $\mathcal{H}$ denotes the model and $\mathcal{F}$ represents the true label function. 
This metric is used to calculate discrepancies on labeled data $\mathcal{D}_L$ and unlabeled data $\mathcal{D}_U$ separately:
\begin{equation}
\footnotesize
\begin{aligned}
    &\xi_{\mathcal{D}_L}(\mathcal{H},\mathcal{F}) = \mathbb{E}_{x \in \mathcal{D}_L} \left[ \xi\left(\mathcal{H}(x), \mathcal{F}(x)\right) \right], \\
    &\xi_{\mathcal{D}_U}(\mathcal{H},\hat{\mathcal{F}}) = \mathbb{E}_{x \in \mathcal{D}_U} \left[ \xi\left(\mathcal{H}(x), \hat{\mathcal{F}}(x)\right) \right],
\end{aligned}
\end{equation}
where $\mathcal{F}$ and $\hat{\mathcal{F}}$ denote the labeling functions for $\mathcal{D}_L$ and $\mathcal{D}_U$, respectively.

\end{definition}

\begin{table}[t]
\footnotesize
    \centering
    \caption{List of symbols for  DIG-FACE.}
    \begin{tabular}{|p{0.25\linewidth}|p{0.65\linewidth}|}
        \hline
        \textbf{Symbol} & \textbf{Description} \\
        \hline
        $\mathbb{R}$ & The model hypothesis space\\
        \hline
        $\mathcal{D}_L$ & Labeled dataset with $\mathcal{N}$ known categories \\
        \hline
        $\mathcal{D}_U = \{\mathcal{D}_U^{\text{old}}, \mathcal{D}_U^{\text{new}}\}$ & Unlabeled dataset with $\mathcal{D}_U^{\text{old}}$ (known, $\mathcal{N}$ categories) and $\mathcal{D}_U^{\text{new}}$ (new, $\mathcal{M}$ categories) \\
        \hline
        $\mathcal{F}$, $\mathcal{\hat{F}}$ & True label space for $\mathcal{D}_L$ and $\mathcal{D}_U$ \\
        \hline
        $\mathcal{H}$ or $\mathcal{H}_{h, \psi} \in \mathbb{R}$ & Main model $\mathcal{H}$ consists of feature exactor $\psi$ and classification head $h$ \\
        \hline
        $\mathcal{H}_{h^*, \psi^*} \in \mathbb{R}$, $\mathcal{H'}_{h_a, \psi_a} \in \mathbb{R}$ & Optimal reference model; auxiliary model for bounding bias \\
        \hline
        $\xi_{\mathcal{D}_L}(\mathcal{H}, \mathcal{F})$, & Discrepancy metrics on $\mathcal{D}_L$ and \\
        $\xi_{\mathcal{D}_U}(\mathcal{H}, \hat{\mathcal{F}})$ & $\mathcal{D}_U$ \\
        \hline
        $\mathbb{P}$, $\mathbb{E}$ & Probability space of model predictions; Euclidean space \\
        \hline
        $f: \mathbb{P} \rightarrow \mathbb{E}$ & Mapping function from probability to Euclidean space \\
        \hline
        $\mathop{\Delta}(\mathcal{D}_U, \mathcal{D}_L)$ & \textit{F-discrepancy}\\
        \hline
    \end{tabular}
    \label{tab:notation}
\end{table}

\begin{figure*}[t]
    \centering
    \includegraphics[width=0.83\linewidth]{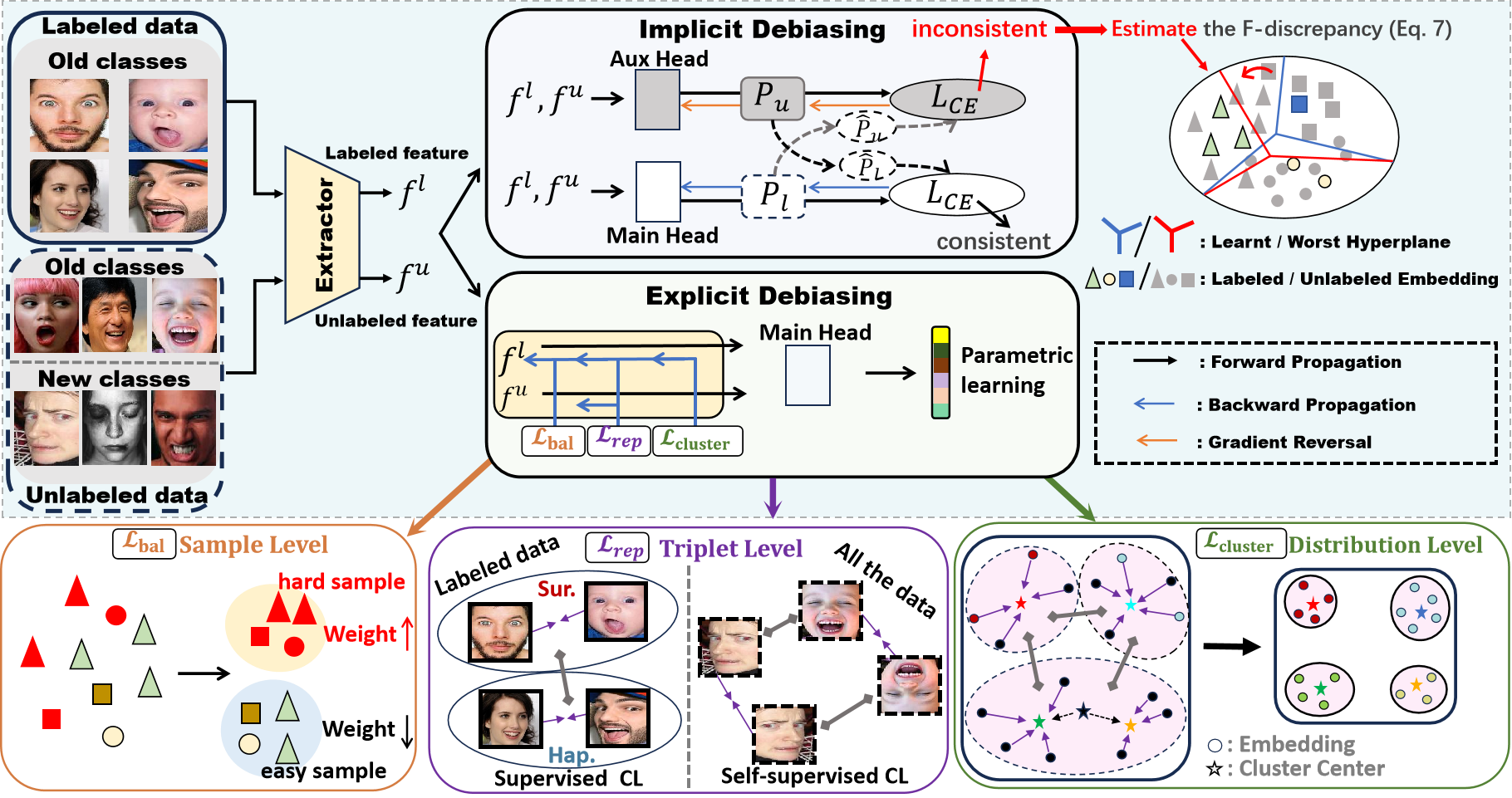}
    \caption{Overview of  \textbf{DIG-FACE} framework. 
    \textbf{(i)} In the implicit debiasing stage, we estimate and minimize the maximum bias by enforcing consistency between the main and auxiliary head on $\mathcal{D}_L$ and inconsistency on  $\mathcal{D}_U$. At this stage, $P_l$ and $P_u$ denote the pseudo-labels of labeled and unlabeled data.
    \textbf{(ii)} In the explicit debiasing stage, we enhance recognition of known and unknown categories through \textbf{sample-level}, \textbf{triplet-level}, and \textbf{distribution-level} optimization, and improve decision-making under G-FACE with parametric learning.
}
    \label{framework}
\end{figure*}

\begin{definition}[Category-based Discrepancy Composition]\label{promotion}
In G-FACE, the unlabeled dataset $\mathcal{D}_U$ contains both new and old categories. Assuming the proportion of old classes follows a probability $\theta$, the discrepancy on $\mathcal{D}_U$ can be decomposed as:
\begin{equation}
\footnotesize
    \xi_{\mathcal{D}_U}(\mathcal{H},\hat{\mathcal{F}}) = (1-\theta)\xi^{new}_{\mathcal{D}_U}(\mathcal{H},\hat{\mathcal{F}}) + \theta\xi^{old}_{\mathcal{D}_U}(\mathcal{H},\hat{\mathcal{F}}),
\end{equation}

where $\xi^{new}_{\mathcal{D}_U}$ and $\xi^{old}_{\mathcal{D}_U}$ denote discrepancies for new and old categories, respectively.
Given $\mathcal{Y}_l \subset \mathcal{Y}_u$, we assume that the model's error on $\mathcal{D}_U$ is at least as large as on $\mathcal{D}_L$:
\begin{equation}
\footnotesize
   \xi_{\mathcal{D}_L}(\mathcal{H},\mathcal{F}) \leq  \xi^{old}_{\mathcal{D}_U}(\mathcal{H},\hat{\mathcal{F}}) \leq \xi_{\mathcal{D}_U}(\mathcal{H},\hat{\mathcal{F}}).
\end{equation}
\end{definition}

\begin{definition}[\textit{F-discrepancy}]\label{bound}  
The \textit{F-discrepancy} is defined as an upper bound on the divergence between discrepancies of $\mathcal{D}_L$ and $\mathcal{D}_U$. It compares the current model \(\mathcal{H}\) with an auxiliary hypothesis \(\mathcal{H'}\) across the hypothesis space \(\mathcal{R}\), and is expressed as:
\begin{equation}
\footnotesize
 \mathop{\Delta}(\mathcal{D}_U,\mathcal{D}_L)  = \sup_{\mathcal{H},\mathcal{H'} \in \mathcal{R}} \left| \xi_{\mathcal{D}_U}(\mathcal{H}, \mathcal{H'}) - \alpha \cdot \xi_{\mathcal{D}_L}(\mathcal{H}, \mathcal{H'}) \right|,
\end{equation}  
where \(\alpha\) adjusts the contribution of labeled data discrepancies.  This definition provides a foundation for bounding implicit bias in subsequent analysis.  
\end{definition}

\begin{lemma}[Upper Bound on Implicit Bias]\label{Lemma}
As the labels of \(\mathcal{D}_U\) are not directly observable, we leverage the labeled data \(\mathcal{D}_L\) to establish an upper bound for the implicit bias caused by new categories. Specifically, the implicit bias satisfies:  
\begin{equation}
\label{bound}
\footnotesize
\xi^{new}_{\mathcal{D}_U}(\mathcal{H},\hat{\mathcal{F}}) \leq \frac{1}{1-\theta} \left[ (\alpha-\theta)\cdot\xi_{\mathcal{D}_L}(\mathcal{H},\mathcal{F}) + \mathop{\Delta}(\mathcal{D}_U,\mathcal{D}_L)+ \lambda \right],
\end{equation}
where $\lambda = \alpha\cdot\xi_{\mathcal{D}_L}(\mathcal{H^*},\mathcal{F})+\xi_{\mathcal{D}_U}(\mathcal{H^*},\mathcal{F})$. The proof is provided in appendix.
\end{lemma} 

\subsection{Explicit Bias}
Compared to implicit bias, explicit bias is more easily observed. It typically manifests as subtle visual features that cause blurry decision boundaries and difficulties in distinguishing categories.
While past FER methods have made significant progress in addressing closed-set~\cite{yu2024exploring,zhang2024leave} and open-set FER~\cite{zhang2024open,liu2024open} that do not require further classification of new classes, 
we identify that G-FACE faces a unique challenge: the subtle differences in the preference on different visual features between known and unknown expressions produce overlapped distributions in the feature spaces, making it difficult to distinguish diverse categories. 
This phenomenon can be partially reflected in the t-SNE visualization of the baseline SimGCD (Fig. \ref{fig:challenge}, bottom), where the overlap in data distributions indicates a clear bias.

\section{DIG-FACE Debiasing Methodology}\label{algorithm}



\subsection{Implicit Debiasing}\label{Constrain}
In Eq. (\ref{bound}) of Lemma \ref{Lemma}, we provide an upper bound for the implicit bias, and thus our goal is to minimize this upper bound to control the learning of new classes.  Specifically, it requires us to leverage the inconsistency between the auxiliary head $\mathcal{H}'$ and the main head $\mathcal{H}$ on  $\mathcal{D}_U$, while ensuring their consistency on $\mathcal{D}_L$. Intuitively, this will enable the feature exactor to maintain robust performance even under poorer decision boundaries, thereby effectively mitigating the impact of decision boundary shifts caused by the \textbf{implicit bias}.
The Min-Max process is as follows:

\begin{equation}\label{minmax}
\footnotesize
\mathop{\min}_{\mathcal{H}}\mathop{\max}_{\mathcal{H'}}  (\alpha-\theta)\cdot\xi_{D_L}(\mathcal{H},\mathcal{F})+ \left| \xi_{\mathcal{D}_U}(\mathcal{H}, \mathcal{H'}) - \alpha\cdot\xi_{\mathcal{D}_L}(\mathcal{H}, \mathcal{H'}) \right|,
\end{equation}
where $(\alpha-\theta)$ is an adjustment parameter, $\psi$ represents the feature exactor. Considering that in \textbf{Definition \ref{Metric}}, we use the mapping function \( f \) to transform the softmax probability vector into Euclidean space, we now apply \( f^{-1}: \mathbb{E} \rightarrow \mathbb{P} \) to revert back to the probability space. Thus, we use cross-entropy  and construct the following adversarial loss:
\begin{equation}\label{L_ad}
\footnotesize
\begin{aligned}
    \mathcal{L}_{ad} =
    \frac{\alpha}{m} \sum_{i=1}^{m} \ell_{CE}&\left( h_a(\psi(x^l_i)), h(\psi(x^l_i)) \right) \\
    - \frac{1}{n} \sum_{j=1}^{n} &\ell_{CE} \left( h_a(\psi(x^u_j)), \hat{\mathcal{P}}^h(x_j^u) \right),
\end{aligned}
\end{equation}

where $x^u$ and $x^l$ are labeled and unlabeled data, respectively, and $\hat{\mathcal{P}^h}$ is the pseudo-label assigned by main classification head $h$ to the unlabeled data, 
$\alpha$ coefficient is set to 2 according to the hyperparameter analysis in the appendix.

\subsection{Explicitly Debiasing }\label{optimize}


Explicit bias arises from overlapping characteristics between known and unknown categories, leading to ambiguous boundaries. To address this, we propose a hierarchical  category-discrimination refinement strategy through \textbf{sample-level}, \textbf{triplet-level}, and \textbf{distribution-level} optimizations, complemented by a parametric classifier to enhance decision-making in the G-FACE task.


\noindent\textbf{Sample-level: Confusing Sample Mining.} To address ambiguity in overlapping characteristics, we propose a dynamic adjustment mechanism applied to reach labeled data $\mathcal{B}^l$. This strategy focuses on mining confusing samples and enhancing clarity in known categories while preserving the model’s ability to discover new categories in unlabeled data. Without requiring prior knowledge of class distribution, the loss function is defined as:



\begin{equation}
\label{balance}
\footnotesize
\begin{aligned}
\mathcal{L}_{\text{bal}}(\mathbf{x}, y; \psi, h)= 
\frac{1}{|\mathcal{B}^l|} \sum_{x_i \in \mathcal{B}^l} \left[ (1 - e_i) + \frac{e_i}{a_{y_i} + \epsilon} \right] \ell_{CE}(h(\psi(\mathbf{x}_i)), y_i),
\end{aligned}
\end{equation}
where \(a_{y_i}\) is the adaptive weight for the true class \(y_i\), calculated as the ratio of \(\varsigma_{y_i}\) (the number of correct predictions for class \(y_i\)) to \(\eta_{y_i}\) (the total number of predictions made for class \(y_i\)), i.e., \(a_{y_i} = \frac{\varsigma_{y_i}}{\eta_{y_i}}\). The time-varying weight \(e_i\) decreases as training progresses, given by 
\(e_i = 0.1 \cdot \left(1 - \frac{t}{T}\right)\), where \(t\) is the current epoch and \(T\) is the epoch. This mechanism dynamically emphasizes underperforming classes, improving the clarity of decision boundaries. Furthermore, for the dynamically adjusted $e_i$, we conduct experimental validation in  the appendix.

\noindent\textbf{Triplet-level: Learning Subtle Facial Expression Features.}  This level focuses on acquiring discriminative features from the feature extractor $\psi$ based on a triplet of samples, enabling the classifier to effectively distinguish between all categories.  To this end, we utilize two types of contrastive learning (CL), including self-supervised CL and supervised CL.  Using any two random augmented versions $\hat{\mathbf{x}}_i$ and $\tilde{\mathbf{x}}_i$ of a face image $\mathbf{x}_i$ within a training batch $\mathcal{B}$ that comprises both labeled and unlabeled samples, the self-supervised CL loss is defined as:
\begin{equation} \footnotesize 
\label{self-supcon}\begin{aligned} \mathcal{L}_{\text{rep}}^{u}(\hat{\mathbf{x}}_i,\tilde{\mathbf{x}}_i;\psi,\tau_u)  &= \frac{1}{|\mathcal{B}|} \sum_{\mathbf{x}_i \in \mathcal{B}} - \log \frac{\exp (\hat{\mathbf{z}}_i^\top \tilde{\mathbf{z}}_i / \tau_u)}{\sum_{\mathbf{x}_k \in \mathcal{B}} \exp (\hat{\mathbf{z}}_k^\top \tilde{\mathbf{z}}_i / \tau_u)}, \end{aligned} \end{equation}
where $\tau_u$ denotes a scaling parameter known as the temperature, $\hat{z}_i$ is the $\ell _2$-normalised vector of $\psi(\hat{\mathbf{x}}_i)$.
Similarly,  
we construct the supervised CL loss~\cite{khosla2020supervised} as $\mathcal{L}_{\text{rep}}^{s}(\hat{\mathbf{x}}_i,\tilde{\mathbf{x}}_i,y;\psi,\tau_c)$ on labeled data. 
These two losses are merged to learn the subtle facial expression representation as: $\mathcal{L}_{\text{rep}} = (1-\lambda) \mathcal{L}_{\text{rep}}^{u} + \lambda \mathcal{L}_{\text{rep}}^{s}$, with $\lambda$ as a hyper-parameter based on previous GCD study~\cite{vaze2022generalized}.\\

\noindent\textbf{Distribution-level: Enhancing Cluster Discriminability.}
To improve the discriminability of expression boundaries, we combine clustering with contrastive learning, which alone often lacks awareness of the global data structure. Specifically, we enhance the feature exactor $\psi$ by incorporating a supervised clustering loss.
The clustering loss $\mathcal{L}_{\text{cluster}}$ comprises two components as $\mathcal{L}_{\text{cluster}} = \mathcal{L}_{\text{WB}} + \beta \cdot \mathcal{L}_{\text{MM}}$, where $\beta$ is the weight of regular term and set as 0.2 in this study. 
The term $\mathcal{L}_{\text{WB}}$ aims to minimize intra-class distances and maximize inter-class distances, while $\mathcal{L}_{\text{MM}}$, as a regularization term, promotes compact clustering by reducing the gap between maximum and minimum distances. Formally, the $\mathcal{L}_{\text{WB}}$ and $\mathcal{L}_{\text{MM}}$ are written as:
\begin{equation}
\footnotesize
\begin{aligned}
&\mathcal{L}_{\text{WB}} = \frac{\sum_{c \in \mathcal{C}^l} \sum_{x_i \in \mathcal{B}^l_c} \|\psi(x_i) - \mu_c\|_2^2}{\sum_{c \in \mathcal{C}^l} n_c \|\mu_c - \mu_g\|_2^2 + \epsilon}, \\
& \mathcal{L}_{\text{MM}}= \frac{1}{|\mathcal{C}^l|} \sum_{c \in \mathcal{C}^l} \left( \max_{x_i \in \mathcal{B}^l_c} \|\psi(x_i) - \mu_c\|_2^2 - \min_{x_i \in \mathcal{B}^l_c} \|\psi(x_i) - \mu_c\|_2^2 \right),
\end{aligned}
\end{equation}
where $\mu_c$ represents the class mean, $\mu_g$ is the global feature mean, and $n_c$ is the number of samples in class $c$. The set $\mathcal{C}^l$ includes all unique class labels in the batch, while $\mathcal{B}^l_c$ denotes the samples belonging to class $c$.  The total $\mathcal{L}_{\text{cluster}}$ is computed per mini-batch of labeled data $\mathcal{B}^l$, activated after $T_{\text{warmup}}$.


\noindent\textbf{Parametric Classifier Learning.}  
The parametric classifier is used to augment the main expression classification head $h$. To enhance the main expression classification head $h$, we employ a parametric classifier following the approach of SimGCD~\cite{wen2023parametric}. Specifically, the number of categories $K=\mathcal{M}+\mathcal{N}$ is given. 
We start by randomly initializing a set of prototypes, one for each category: $\mathcal{T} = \{\mathbf{t}_1, \mathbf{t}_2, \dots, \mathbf{t}_K\}$ is randomly initialized at the beginning. During training, the soft label $\hat{\mathbf{p}}_i^k$ for each augmented view $\mathbf{x}_i$ is calculated using a softmax function on the cosine similarity between the hidden feature and the prototypes:
\begin{equation}
\small
    \hat{\mathbf{p}}_i^k = \frac{\exp\left(\frac{1}{\tau_s}(h(\psi(\mathbf{x}_i)) / {\|h(\psi(\mathbf{x}_i))\|_2})^\top ({\mathbf{t}_k}/{\|\mathbf{t}_k\|_2})\right)}{\sum_{j}\exp{\left(\frac{1}{\tau_s} (h(\psi(\mathbf{x}_i))/\|h(\psi(\mathbf{x}_i))\|_2)^\top (\mathbf{t}_j/\|\mathbf{t}_j\|_2)\right)}}.
\end{equation}
Similarly, we can obtain the soft label $\tilde{\mathbf{p}}_i$ of the view $\tilde{\mathbf{x}}_i$.
Then the objective of parametric classifier learning is the sum of the supervised loss and unsupervised loss as  $\mathcal{L}_{\text{cls}}=(1-\lambda) \mathcal{L}_{\text{cls}}^u+\lambda \mathcal{L}_{\text{cls}}^s$.  The supervised and unsupervised losses are formulated by:
\begin{equation}
\footnotesize
\begin{aligned}\label{Cls loss}
&\mathcal{L}_{\text{cls}}^s = \frac{1}{|\mathcal{B}^l|}\sum_{\mathbf{x}_i \in \mathcal{B}^l} \ell_{CE}(\mathbf{y}({x_i}), \hat{\mathbf{p}}_i), \,\\
&\mathcal{L}_{\text{cls}}^u = \frac{1}{|\mathcal{B}|}\sum_{\mathbf{x}_i \in \mathcal{B}} \ell_{CE}(\tilde{\mathbf{p}}_i, \hat{\mathbf{p}}_i) - \epsilon H(\overline{\mathbf{p}}),
\end{aligned}
\end{equation}
where $\mathcal{B}^l$ is the mini-batch of labeled training data, $\mathbf{y}({x_i})$ is the ground truth label for the labeled data point $\mathbf{x}_i$, $\ell_{CE}$ is the cross-entropy loss, and $H(\overline{\mathbf{p}})=-\sum \overline{\mathbf{p}} \log \overline{\mathbf{p}}$ regularizes the mean prediction $\overline{\mathbf{p}}=\frac{1}{2|\mathcal{B}|}\sum_{\mathbf{x}_i \in \mathcal{B}}(\hat{\mathbf{p}}_i + \tilde{\mathbf{p}}_i)$ in a mini-batch.

\subsection{Overall Loss Function:}
The overall loss function in our  DIG-FACE is as follows:
\begin{equation}
    \mathcal{L}_{total} = \mathcal{L}_{rep} + \mathcal{L}_{cls} + \lambda_a\mathcal{L}_{ad} +  \lambda_b\mathcal{L}_{bal} + 
 \lambda_c\mathcal{L}_{cluster} ,
\end{equation}
where $\lambda_a$, $\lambda_b$ and $\lambda_c$ are balancing parameters for the contributions of two debasing processes during training, we set them to 0.2, 0.3, and 0.2, respectively, based on the hyperparameter analysis in our supplementary appendix. $\mathcal{L}_{cluster}$ is only introduced after the warmup epoch $T_\text{warmup}$, when the model has preliminary clustering capabilities.  

\begin{table*}[h]
\caption{\textbf{Category discovery accuracy (ACC) on RAF-DB dataset. Above the table are the four selected known categories.}}
\centering
\footnotesize
\begin{tabular}{@{}lccccccccc@{}}
\toprule
& \multicolumn{3}{c}{Sur.+Fea.+Dis.+Hap.} & \multicolumn{3}{c}{Sur.+Fea.+Dis.+Ang.} & \multicolumn{3}{c}{Sur.+Hap.+Sad.+Neu.} \\ \cmidrule(l){2-4} \cmidrule(l){5-7} \cmidrule(l){8-10}
Method       & All     & Old     & New     & All     & Old     & New     & All     & Old     & New     \\ \midrule
k-means~\cite{macqueen1967some}         & 27.8 & 28.5 & 27 & 26.2 & 23.9 & 26.4 & 26.2 & 25.7 & 27.5 \\
GCD~\cite{vaze2022generalized}\footnotesize{\textcolor{gray}{[CVPR'22]}}          & 31.2 & 33.3 & 28.5 & 33.2 & 36.8 & 32.6 & 56.8 & 63.5 & 36.1 \\
GPC~\cite{zhao2023learning}\footnotesize{\textcolor{gray}{[ICCV'23]}}         & 47.2 &56.7& 38.4 & 45.5  &  47.3 & 43.2 & 59.4 & 67.1 & 38.4\\
InfoSieve~\cite{rastegar2024learn}\footnotesize{\textcolor{gray}{[NeurIPS'23]}}    & \underline{52.8} & \textbf{72.2} & 40.1 & 50.1 & 62.2 & 48.3 & 58.2 & 64.6 & 39.6 \\
CMS-GCD~\cite{choi2024contrastive}\footnotesize{\textcolor{gray}{[CVPR'24]}}    & 46.6 & 55.6 & \underline{40.5} & 50.8 & 63.5 & 49.7 & \underline{64.0} & \underline{74.5} & 38.5 \\
\hline
SimGCD~\cite{wen2023parametric}\footnotesize{\textcolor{gray}{[ICCV'23]}}       & 43.3 & 47.9& 40.2 & 49.6 & 56.9 & 48.5 & 57.1 & 62.1 & \underline{41.7} \\
DIG-FACE        & \textbf{53.3} &\underline{71.3}& \textbf{41.0} & \textbf{52.4} &  \textbf{66.3} & \textbf{50.2} & \textbf{69.1}&  \textbf{77.8} & \textbf{41.9}\\
\bottomrule
\end{tabular}
\label{rafdb basic}
\end{table*}

\begin{table*}[h]
\caption{\textbf{Category discovery accuracy (ACC) on FerPlus and AffectNet datasets.}}
\centering
\scriptsize 
\begin{tabular}{@{}lcccccccccccccccccc@{}}
\toprule
& \multicolumn{9}{c}{\textbf{FerPlus}} & \multicolumn{9}{c}{\textbf{AffectNet}} \\ \cmidrule(l){2-10} \cmidrule(l){11-19} 
& \multicolumn{3}{c}{Sur.+Fea.+Dis.+Hap.} & \multicolumn{3}{c}{Sur.+Fea.+Dis.+Ang.} & \multicolumn{3}{c}{Sur.+Hap.+Sad.+Neu.} & \multicolumn{3}{c}{Sur.+Fea.+Dis.+Hap.} & \multicolumn{3}{c}{Sur.+Fea.+Dis.+Ang.} & \multicolumn{3}{c}{Sur.+Hap.+Sad.+Neu.} \\ \cmidrule(l){2-4} \cmidrule(l){5-7} \cmidrule(l){8-10} \cmidrule(l){11-13} \cmidrule(l){14-16} \cmidrule(l){17-19}
Method       & All     & Old     & New     & All     & Old     & New     & All     & Old     & New     & All     & Old     & New     & All     & Old     & New     & All     & Old     & New     \\ \midrule
k-means~\cite{macqueen1967some} & 20.8 & 22.1 & 20.2 & 22.5 & 23.0 & 22.3 & 25.1 & 25.6 & 24.0 & 14.4 & 14.6 & 14.3 & 15.0 & 20.3 & 13.2 & 14.2 & 15.2 & 13.8 \\
\hline
GCD~\cite{vaze2022generalized} & 40.8 & 62.3 & 27.4 & 40.4 & 52.2 & 32.2 & 37.2 & 50.4 & 29.7 & 35.0 & 47.5 & 28.1 & 32.2 & 42.0 & 28.9 & 28.0 & 38.2 & 24.6 \\
GPC~\cite{zhao2023learning} & 45.2 & 63.7 & 34.4 & 45.7 & 64.3 & 40.0 & 52.2 & 58.5 & 46.8 & 36.3 & 46.5 & 33.4 & 38.7 & 50.5 & 32.1 & 33.6 & 44.1 & 30.0 \\
InfoSieve~\cite{rastegar2024learn} & 48.2 & 74.1 & 37.7 & \underline{47.5} & \underline{72.5} & \textbf{44.1} & \underline{64.1} & 69.5 & 53.7 & \underline{47.1} & \underline{54.2} & \underline{35.7} & \underline{49.8} & \underline{68.4} & \underline{37.2} & \underline{46.2} & \underline{57.3} & \textbf{31.4} \\
CMS-GCD~\cite{choi2024contrastive} & \underline{48.9} & \underline{77.6}& 36.5& 46.5 & 78.5 & 40.1 & 62.7& \textbf{75.2} & 47.8 & - & - & - & - & - & -& -& - & - \\
\hline
SimGCD~\cite{wen2023parametric}  & 45.8 & 65.4 & \underline{38.3} & 46.4 & 70.2 & 42.2 & 63.7 & 67.2 & \underline{55.0} & 45.6 & 53.4 & 35.1 & 40.7 & 55.1 & 34.5 & 38.1 & 48.1 & 29.7 \\
DIG-FACE & \textbf{51.8} & \textbf{84.9} & \textbf{38.8} & \textbf{48.6} & \textbf{81.7} & \underline{43.5} & \textbf{68.6} & \underline{74.3} & \textbf{56.4} & \textbf{57.7} & \textbf{62.9} & \textbf{40.9} & \textbf{54.2} & \textbf{74.8} & \textbf{39.4} & \textbf{55.9} & \textbf{69.4} & \underline{30.8} \\
\bottomrule
\end{tabular}
\label{tab:combined_results}
\end{table*}

\section{Experiments}
\subsection{Experimental Setup}
\noindent\textbf{Dataset.} We conduct experiments on three popular FER datasets, which have been partitioned in the format suitable for G-FACE.  \textbf{RAF-DB} ~\cite{li2017reliable} contains eleven compound expressions and seven basic expressions. For G-FACE, we treat four out of the seven basic expressions as known categories, with three different selection methods. Notably, in the stage of discovering compound expressions, we use the basic expressions as known categories. \textbf{FERPlus}~\cite{barsoum2016training} is extended from FER2013~\cite{goodfellow2013challenges}. We used eight categories and selected four as known categories to discover the remaining new categories.
\textbf{AffectNet}~\cite{mollahosseini2017affectnet} is a large-scale FER dataset.  We used eight categories and selected four as known categories to discover the remaining new categories. A summary of dataset statistics  is shown in the appendix. Please refer to the appendix for a statistical summary of the dataset and the reasons for the specific category partitioning.\\
\noindent\textbf{Evaluation Metric.}\label{metric}
Following GCD~\cite{vaze2022generalized} guidelines, we evaluate model performance using Category Discovery Accuracy (ACC). The Hungarian algorithm~\cite{wright1990speeding} is used to optimally assign emerged clusters to their ground truth labels.  The ACC formula is given by $\text{ACC} = \frac{1}{Z} \sum_{i=1}^{Z} \mathbb{I}(y^t_i = q(\hat{y}_i))$, where $Z = |\mathcal{D}_U|$, representing the total number of samples in the unlabeled dataset, and $q$ is the optimal permutation that best matches the predicted cluster assignments to the ground truth labels. We report the accuracies of all the classes (``All''), old classes (``Old'') and unseen classes (``New'').\\
\noindent\textbf{Implementation details.}
For a fair comparison, we conducted the comparison methods five times on each dataset, and ultimately reported the set of results where the "All" accuracy ranked third. We employed a ViT-B/16 backbone (feature exactor $\psi$) network~\cite{dosovitskiy2020image} pre-trained with DINO~\cite{caron2021emerging} (decision head). Training was performed using an initial learning rate of 0.1, which was decayed with a cosine annealed schedule~\cite{loshchilov2016sgdr}. The max training epoch $T$ is set to 200 and batch size of 128 for training. We followed SimGCD~\cite{wen2023parametric} to set the balancing factor $\lambda$ to 0.35, and the temperature
values $\tau_c$ and $\tau_u$ to 0.1 and 0.07, respectively. We initially set $\tau_t$ to 0.07 and $\tau_s$ to 0.1 for the classification objective. Then, a cosine schedule was employed to gradually reduce $\tau_t$ to 0.04 over the first 30 epochs. $\lambda_a$, $\lambda_b$, and $\lambda_c$ were assigned values of 0.2, 0.3, and 0.2, respectively. $T_{warmup}$ was set to 50 epochs. The coefficients $\alpha$ and $\beta$ were set to 0.2 and 0.1, respectively.
The All experiments were conducted using an NVIDIA GeForce RTX 3090 GPU.

\begin{table}[t]
\caption{\textbf{Category discovery accuracy (ACC) on RAF-DB-Compound.} We utilize the seven basic expressions training to discover new compound expressions.}
\footnotesize
\centering
\begin{tabular}{@{}lccc@{}}
\toprule
Method       & All     & Old     & New     \\ 
\midrule
k-means~\cite{macqueen1967some} & 15.7   & 16.6    & 13.9    \\
\hline
GCD~\cite{vaze2022generalized}\footnotesize{\textcolor{gray}{[CVPR'22]}}  & 22.0   & 26.4    & 20.1    \\
GPC~\cite{zhao2023learning}\footnotesize{\textcolor{gray}{[ICCV'23]}}   & 25.2     & 28.1     & 22.3    \\
InfoSieve~\cite{rastegar2024learn}\footnotesize{\textcolor{gray}{[NeurIPS'23]}}   & \underline{40.1}   & \underline{50.5}     & 24.2     \\
\hline
SimGCD~\cite{wen2023parametric}\footnotesize{\textcolor{gray}{[ICCV'23]}}  & 32.6   & 39.0    & \underline{24.8}    \\
\bottomrule
\end{tabular}
\label{compound}
\end{table}

\subsection{Comparison with the State-of-the-Art methods}

\textbf{Comparision with GCD Methods.} 
The effectiveness of DIG-FACE debiasing is most directly reflected in its ability to maintain high accuracy for old facial expression categories while also achieving excellent results in discovering new facial expression categories.
In Tab. \ref{rafdb basic}, we present the comparison with the SOTA GCD method on the basic classes of the RAF-DB dataset ~\cite{li2017reliable}.
In terms of old ACC, we achieved an average improvement of 16.2\% compared to SimGCD~\cite{wen2023parametric}. And the ability to discover new expressions has increased steadily.  Moreover, Tab. \ref{tab:combined_results} shows the results on the dataset FerPlus~\cite{barsoum2016training}. 
Although InfoSeive~\cite{rastegar2024learn} shows marginal improvement in the old class in a certain division (\textit{i.e.}, Surprise, Fear, Disgust, and Anger), our method still achieves significant accuracy gains across all classes, including the new class, demonstrating its effective elimination of implicit bias in learning. 
Furthermore, Tab. \ref{tab:combined_results} shows performance comparison on the large-scale dataset AffectNet~\cite{mollahosseini2017affectnet}. On this challenging dataset, we both achieved the best new category, old category accuracy. Notably, Due to the excessive number of samples in AffectNet, CMS-GCD~\cite{choi2024contrastive} struggled to produce effective results. 
In Tab. \ref{compound}, we report the comparison on the challenging RAF-DB-Compound dataset that discovers composite expressions based on basic expressions, we achieved an improvement of 20.8\% accuracy for the old categories and 4.9\% for the new categories, compared to SimGCD.
In summary, our approach effectively eliminates the bias towards new categories while maintaining high accuracy for old categories, and steadily improves the recognition accuracy for new categories. \\

\noindent\textbf{Comparision with FER Method.}  
To verify the generalizability of our method over previous FER approaches, we selected Ada-CAM~\cite{li2022towards}, a semi-supervised method adapted for the G-FACE task, as a baseline. Ada-CAM achieves an average \textbf{All ACC} of only 10.5\%. By integrating our DIG-FACE framework into Ada-CAM, we achieve notable improvements of 5.7\%, 4.8\%, and 2.1\% in \textbf{All ACC} across different settings, demonstrating the superior adaptability of our approach. Detailed results are provided in the supplemental appendix.
Additionally, adapting to the recent Open-set FER scenario introduced by \citet{zhang2024open} \textit{et al.}, we used AUROC (higher is better) and FPR@TPR95 (lower is better) as metrics. The figure is shown  in the appendix, 
DIG-FACE achieved higher AUROC, indicating strong performance in detecting new categories, and lower FPR@TPR95, reflecting greater decision confidence.

\begin{table}[t]
\centering
\caption{Ablation studies of implicit and explicit debiasing on RAF-DB, with the four known classes shown above.}
\footnotesize
\begin{tabular}{@{}cc|ccc|ccc@{}}
\toprule
\multicolumn{2}{c|}{\textbf{Components}} & \multicolumn{3}{c|}{Sur.+Fea.+Dis.+Hap.} & \multicolumn{3}{c}{Sur.+Fea.+Dis.+Ang.} \\ 
\cmidrule(l){1-8} 
\textbf{Implicit} & \textbf{Explicit} & \textbf{All} & \textbf{Old} & \textbf{New} & \textbf{All} & \textbf{Old} & \textbf{New} \\ 
\midrule
& & 43.3 & 47.9 & 40.2 & 49.6 & 56.9 & 48.5 \\
$\checkmark$ & & 50.4 & 63.2 & 40.5 & 50.5 & 60.5 & 48.6 \\
& $\checkmark$ & 49.8 & 61.1 & 40.7 & 51.5 & 63.2 & 50.3 \\
$\checkmark$ & $\checkmark$ &\textbf{ 53.3} &\textbf{ 71.3} & \textbf{41.0} & \textbf{52.4} & \textbf{66.3} & \textbf{50.2} \\ 
\bottomrule
\end{tabular}
\label{ablation_v1}
\end{table}

\begin{figure}[h]
    \centering
\includegraphics[width=0.95\linewidth]{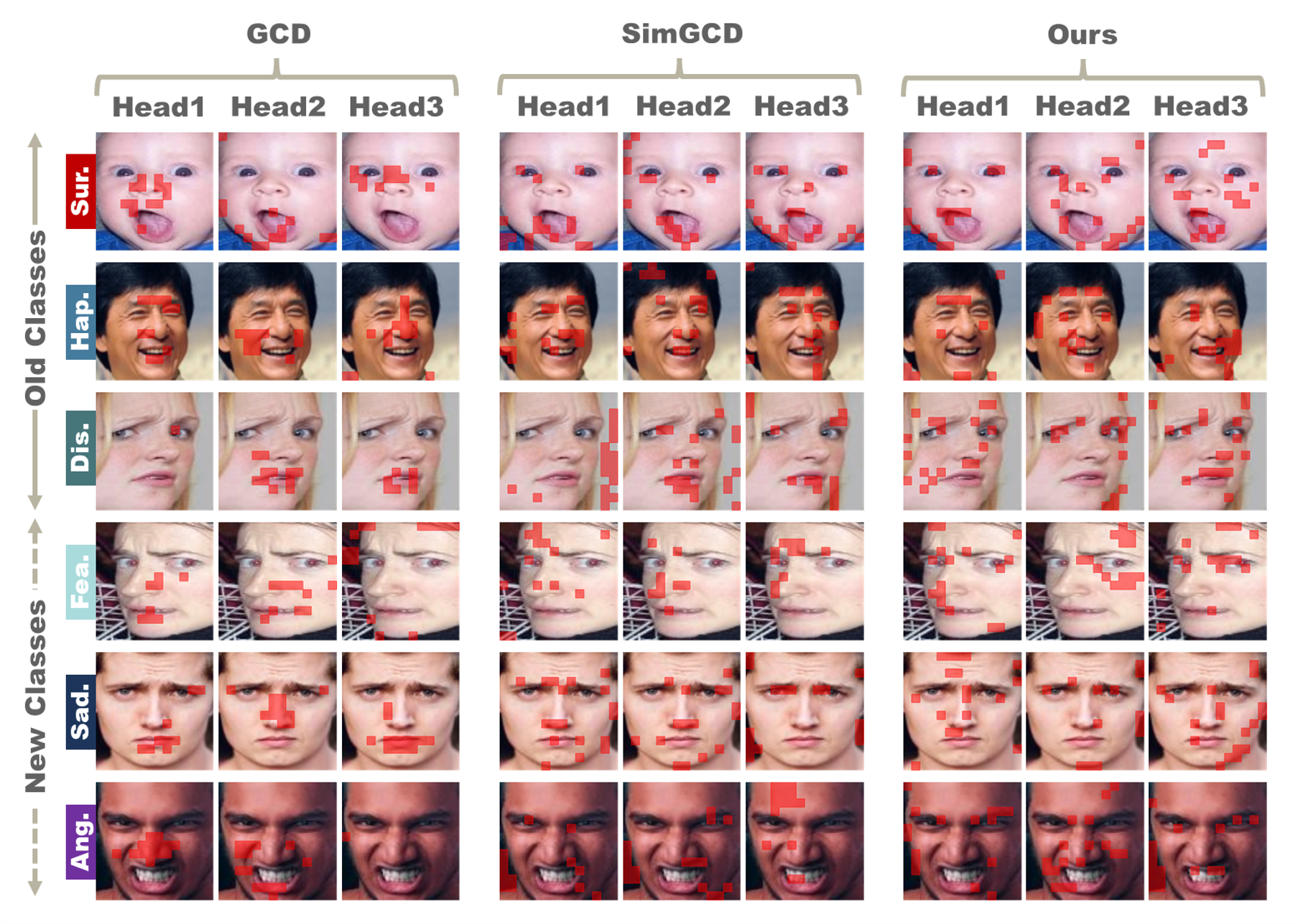}
    \caption{Attention visualization of different self-attention heads (numbered as h1 to h3) on RAF-DB. The top 10\% attended patches are shown in red. Our method pays more attention to the cheeks, eyes and mouth corners details.}
    \label{attention-map}
\end{figure}

\subsection{Ablation Studies} 
To verify the effectiveness of two debiasing components in \textbf{DIG-FACE}, we conducted ablation experiments on the RAF-DB dataset. The results are presented in Tab. \ref{ablation_v1} and the more details table in the appendix.

\noindent\textbf{Implicit Debiasing} is designed to ensure that learning for known categories remains unaffected by the introduction of new categories. Adding this component to the baseline SimGCD in Tab. \ref{ablation_v1} resulted in absolute improvements of 15.3\% and 3.6\%  for known categories and relative improvements of 0.75\% and 0.21\% for new categories. This component focuses on minimizing the bias upper bound, ensuring robust feature extraction even with sub-optimal decision boundaries, thereby bridging the gap between old and new category learning.  


\noindent\textbf{Explicit Debiasing.} Our explicit debiasing component improves feature separability between known and new categories through \textbf{Sample-Level} and\textbf{ Distribution-Level} optimizations, with \textbf{Triplet-Level} contrastive learning adopted from prior GCD work. Incorporating the entire explicit debiasing module leads to absolute improvements of 13.2\% and 6.3\% for known categories and relative improvements of 1.2\% and 3.7\% for new categories. Notably, combining the sample-level optimization with implicit debiasing achieves All ACC improvements of 9.0\% and 1.1\%, while combining the distribution-level optimization with implicit debiasing results in All ACC improvements of 9.8\% and 1.3\%. Further ablation studies are provided in the \textbf{appendix}.\\
\noindent \textbf{In general}, the combination of implicit and explicit debiasing components leads to the most significant performance improvements. Together, they achieve absolute gains of 23.4\% and 9.4\% for known categories and relative improvements of 2.0\% and 3.5\% for new categories, as shown in Tab.\ref{ablation_v1}. This demonstrates that implicit debiasing stabilizes learning for known categories, while explicit debiasing enhances feature separability, resulting in complementary benefits for overall model performance.


\begin{figure}[t]
    \centering
    \includegraphics[width=1\linewidth]{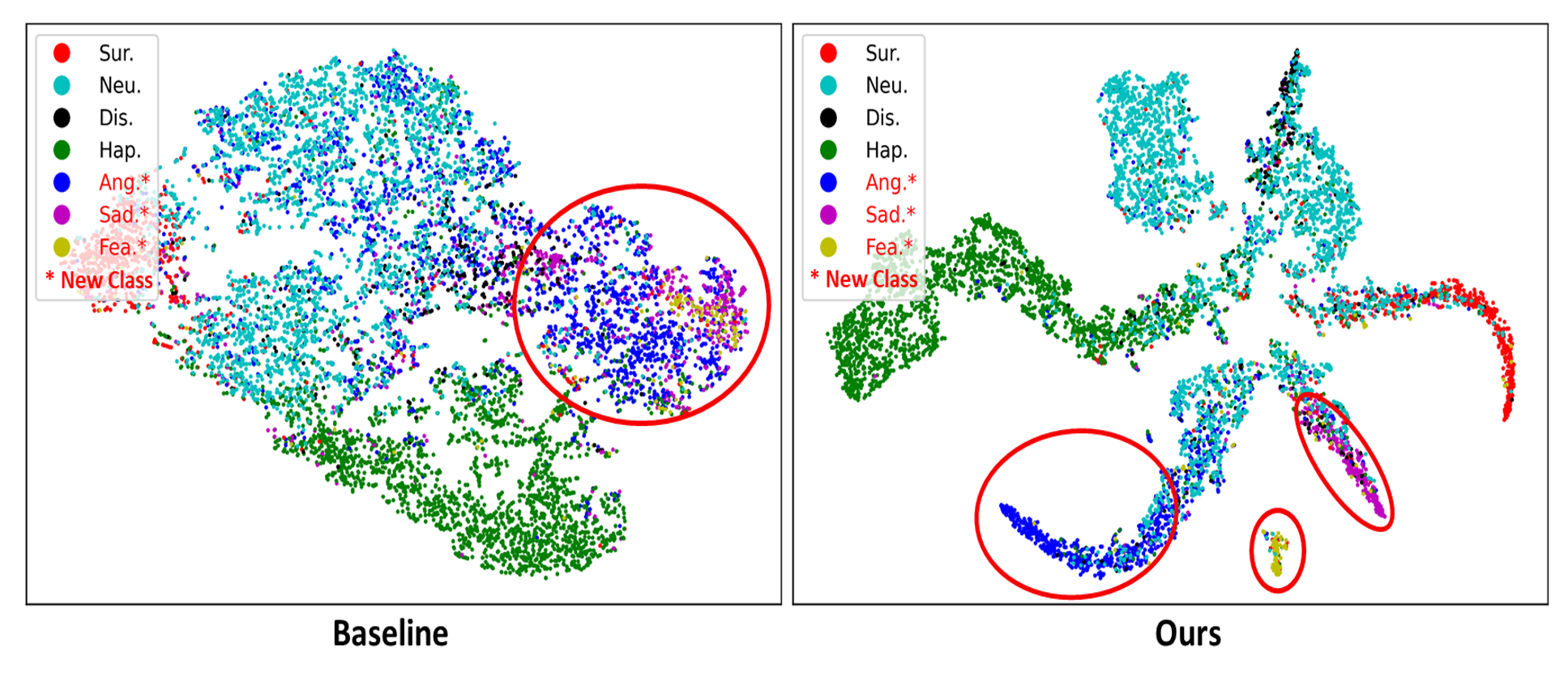}
    \caption{Compared to the baseline SimGCD, the features extracted by our method are more separable in t-SNE visualization.}
    \label{fig:t-SNE}
\end{figure}

\begin{table}[t]
\caption{Results on two general image recognition tasks, namely Herbarium 19 and CIFAR10.}
\footnotesize
\centering
\begin{tabular}{@{}lcccccc@{}}
\toprule
& \multicolumn{3}{c}{Herbarium 19} & \multicolumn{3}{c}{CIFAR10} \\ \cmidrule(l){2-4} \cmidrule(l){5-7}
Methods       & All     & Old     & New     & All     & Old     & New     \\ \midrule
$k$-means~\cite{macqueen1967some} & 13.0 & 12.2 & 13.4 & 83.6 &85.7 &82.5 \\
UNO+~\cite{fini2021unified} & 28.3 & 53.7 & 14.7 & 68.6& \textbf{98.3} &53.8 \\
\hline
GCD~\cite{vaze2022generalized} & 35.4 & 51.0 & 27.0 & 91.5 &\underline{97.9}& 88.2 \\
CMS-GCD~\cite{choi2024contrastive} & 36.4 & 54.9 & 26.4 & 94.5 &96.0& 93.2\\
SimGCD~\cite{wen2023parametric} & \underline{44.0} &\underline{58.0} &\underline{36.4} & 97.1 &95.1 &\textbf{98.1}  \\
DIG-FACE & \textbf{45.2} & \textbf{58.7} & \textbf{37.9} & 97.0 & 95.3 & \underline{97.8} \\
\bottomrule
\end{tabular}
\label{general}
\end{table}
\section{Analysis}


\subsection{Visualization}
As shown in Fig. \ref{attention-map}, we adopt the method from GCD~\cite{vaze2022generalized} to analyze the attention maps generated by the DINO-ViT model. Specifically, we visualize Heads 1 through 3, the self-attention heads in ViT, which reveal how the model captures subtle facial expression details. Compared to GCD and SimGCD, DIG-FACE places greater emphasis on key regions like the cheeks and mouth corners, providing evidence for the improved performance of DIG-FACE.
Besides, the t-SNE visualization results shown in Fig. \ref{fig:t-SNE} demonstrate a more discrete feature representation learned by DIG-FACE, compared to the more blurred feature boundaries via SimGCD. It shows that DIG-FACE can better reduce the overlap of feature spaces between known and unknown expression classes.

\subsection{Performance on General Datasets}

In addition to GCD-based FER datasets, we evaluate DIG-FACE on two general image classification datasets, namely CIFAR-10 \cite{krizhevsky2009learning} and Herbarium 19 \cite{tan2019herbarium}, 
to assess its generalization ability. 
Compared to prior GCD methods and the SimGCD baseline, DIG-FACE demonstrates superior robustness and adaptability, as shown in Tab .\ref{general}. 

\section{Conclusion}
In this paper, we introduced\textbf{ G-FACE}, a novel task addressing the dual challenge of recognizing known facial expressions and discovering unseen ones. We defined two key challenges in this task: \textbf{implicit bias}, arising from distributional gaps between labeled and unlabeled data, and \textbf{explicit bias}, caused by shifts in visual facial characteristics between known and unknown categories. To address these challenges, we proposed DIG-FACE, a de-biasing framework that includes implicit debiasing, which minimizes bias through adversarial learning, and explicit debiasing, which enhances feature separability to address category ambiguity. Extensive experiments demonstrate its effectiveness, significantly enhancing recognition for both known and new categories, setting a strong benchmark for G-FACE.

{
    \small
    \bibliographystyle{ieeenat_fullname}
    \bibliography{refs}

\begin{thebibliography}{49}
\providecommand{\natexlab}[1]{#1}
\providecommand{\url}[1]{\texttt{#1}}
\expandafter\ifx\csname urlstyle\endcsname\relax
  \providecommand{\doi}[1]{doi: #1}\else
  \providecommand{\doi}{doi: \begingroup \urlstyle{rm}\Url}\fi

\bibitem[Banerjee et~al.(2024)Banerjee, Kallooriyakath, and Biswas]{banerjee2024amend}
Anwesha Banerjee, Liyana~Sahir Kallooriyakath, and Soma Biswas.
\newblock Amend: Adaptive margin and expanded neighborhood for efficient generalized category discovery.
\newblock In \emph{Proceedings of the IEEE/CVF Winter Conference on Applications of Computer Vision}, pages 2101--2110, 2024.

\bibitem[Barsoum et~al.(2016)Barsoum, Zhang, Ferrer, and Zhang]{barsoum2016training}
Emad Barsoum, Cha Zhang, Cristian~Canton Ferrer, and Zhengyou Zhang.
\newblock Training deep networks for facial expression recognition with crowd-sourced label distribution.
\newblock In \emph{Proceedings of the 18th ACM international conference on multimodal interaction}, pages 279--283, 2016.

\bibitem[Calder and Young(2016)]{calder2016understanding}
Andrew~J Calder and Andrew~W Young.
\newblock Understanding the recognition of facial identity and facial expression.
\newblock \emph{Facial Expression Recognition}, pages 41--64, 2016.

\bibitem[Caron et~al.(2021)Caron, Touvron, Misra, J{\'e}gou, Mairal, Bojanowski, and Joulin]{caron2021emerging}
Mathilde Caron, Hugo Touvron, Ishan Misra, Herv{\'e} J{\'e}gou, Julien Mairal, Piotr Bojanowski, and Armand Joulin.
\newblock Emerging properties in self-supervised vision transformers.
\newblock In \emph{Proceedings of the IEEE/CVF international conference on computer vision}, pages 9650--9660, 2021.

\bibitem[Choi et~al.(2024)Choi, Kang, and Cho]{choi2024contrastive}
Sua Choi, Dahyun Kang, and Minsu Cho.
\newblock Contrastive mean-shift learning for generalized category discovery.
\newblock In \emph{Proceedings of the IEEE/CVF Conference on Computer Vision and Pattern Recognition}, pages 23094--23104, 2024.

\bibitem[Cowen et~al.(2021)Cowen, Keltner, Schroff, Jou, Adam, and Prasad]{cowen2021sixteen}
Alan~S Cowen, Dacher Keltner, Florian Schroff, Brendan Jou, Hartwig Adam, and Gautam Prasad.
\newblock Sixteen facial expressions occur in similar contexts worldwide.
\newblock \emph{Nature}, 589\penalty0 (7841):\penalty0 251--257, 2021.

\bibitem[Dosovitskiy et~al.(2020)Dosovitskiy, Beyer, Kolesnikov, Weissenborn, Zhai, Unterthiner, Dehghani, Minderer, Heigold, Gelly, et~al.]{dosovitskiy2020image}
Alexey Dosovitskiy, Lucas Beyer, Alexander Kolesnikov, Dirk Weissenborn, Xiaohua Zhai, Thomas Unterthiner, Mostafa Dehghani, Matthias Minderer, Georg Heigold, Sylvain Gelly, et~al.
\newblock An image is worth 16x16 words: Transformers for image recognition at scale.
\newblock \emph{arXiv preprint arXiv:2010.11929}, 2020.

\bibitem[Ekman(1993)]{ekman1993facial}
Paul Ekman.
\newblock Facial expression and emotion.
\newblock \emph{American psychologist}, 48\penalty0 (4):\penalty0 384, 1993.

\bibitem[Fei et~al.(2022)Fei, Zhao, Yang, and Zhao]{fei2022xcon}
Yixin Fei, Zhongkai Zhao, Siwei Yang, and Bingchen Zhao.
\newblock Xcon: Learning with experts for fine-grained category discovery.
\newblock \emph{arXiv preprint arXiv:2208.01898}, 2022.

\bibitem[Fini et~al.(2021)Fini, Sangineto, Lathuili{\`e}re, Zhong, Nabi, and Ricci]{fini2021unified}
Enrico Fini, Enver Sangineto, St{\'e}phane Lathuili{\`e}re, Zhun Zhong, Moin Nabi, and Elisa Ricci.
\newblock A unified objective for novel class discovery.
\newblock In \emph{Proceedings of the IEEE/CVF International Conference on Computer Vision}, pages 9284--9292, 2021.

\bibitem[Goodfellow et~al.(2013)Goodfellow, Erhan, Carrier, Courville, Mirza, Hamner, Cukierski, Tang, Thaler, Lee, et~al.]{goodfellow2013challenges}
Ian~J Goodfellow, Dumitru Erhan, Pierre~Luc Carrier, Aaron Courville, Mehdi Mirza, Ben Hamner, Will Cukierski, Yichuan Tang, David Thaler, Dong-Hyun Lee, et~al.
\newblock Challenges in representation learning: A report on three machine learning contests.
\newblock In \emph{Neural Information Processing: 20th International Conference, ICONIP 2013, Daegu, Korea, November 3-7, 2013. Proceedings, Part III 20}, pages 117--124. Springer, 2013.

\bibitem[Han et~al.(2019)Han, Vedaldi, and Zisserman]{han2019learning}
Kai Han, Andrea Vedaldi, and Andrew Zisserman.
\newblock Learning to discover novel visual categories via deep transfer clustering.
\newblock In \emph{Proceedings of the IEEE/CVF International Conference on Computer Vision}, pages 8401--8409, 2019.

\bibitem[Han et~al.(2021)Han, Rebuffi, Ehrhardt, Vedaldi, and Zisserman]{han2021autonovel}
Kai Han, Sylvestre-Alvise Rebuffi, Sebastien Ehrhardt, Andrea Vedaldi, and Andrew Zisserman.
\newblock Autonovel: Automatically discovering and learning novel visual categories.
\newblock \emph{IEEE Transactions on Pattern Analysis and Machine Intelligence}, 44\penalty0 (10):\penalty0 6767--6781, 2021.

\bibitem[Kanade et~al.(2000)Kanade, Cohn, and Tian]{kanade2000comprehensive}
Takeo Kanade, Jeffrey~F Cohn, and Yingli Tian.
\newblock Comprehensive database for facial expression analysis.
\newblock In \emph{Proceedings fourth IEEE international conference on automatic face and gesture recognition (cat. No. PR00580)}, pages 46--53. IEEE, 2000.

\bibitem[Khosla et~al.(2020)Khosla, Teterwak, Wang, Sarna, Tian, Isola, Maschinot, Liu, and Krishnan]{khosla2020supervised}
Prannay Khosla, Piotr Teterwak, Chen Wang, Aaron Sarna, Yonglong Tian, Phillip Isola, Aaron Maschinot, Ce Liu, and Dilip Krishnan.
\newblock Supervised contrastive learning.
\newblock \emph{Advances in neural information processing systems}, 33:\penalty0 18661--18673, 2020.

\bibitem[Kollias(2023)]{Kollias_2023_CVPR}
Dimitrios Kollias.
\newblock Multi-label compound expression recognition: C-expr database \& network.
\newblock In \emph{Proceedings of the IEEE/CVF Conference on Computer Vision and Pattern Recognition (CVPR)}, pages 5589--5598, 2023.

\bibitem[Krizhevsky et~al.(2009)Krizhevsky, Hinton, et~al.]{krizhevsky2009learning}
Alex Krizhevsky, Geoffrey Hinton, et~al.
\newblock Learning multiple layers of features from tiny images.
\newblock 2009.

\bibitem[Kumari et~al.(2015)Kumari, Rajesh, and Pooja]{kumari2015facial}
Jyoti Kumari, Reghunadhan Rajesh, and KM Pooja.
\newblock Facial expression recognition: A survey.
\newblock \emph{Procedia computer science}, 58:\penalty0 486--491, 2015.

\bibitem[Li et~al.(2022)Li, Wang, Yang, Wang, and Gao]{li2022towards}
Hangyu Li, Nannan Wang, Xi Yang, Xiaoyu Wang, and Xinbo Gao.
\newblock Towards semi-supervised deep facial expression recognition with an adaptive confidence margin.
\newblock In \emph{Proceedings of the IEEE/CVF conference on computer vision and pattern recognition}, pages 4166--4175, 2022.

\bibitem[Li and Deng(2020)]{li2020deep}
Shan Li and Weihong Deng.
\newblock Deep facial expression recognition: A survey.
\newblock \emph{IEEE transactions on affective computing}, 13\penalty0 (3):\penalty0 1195--1215, 2020.

\bibitem[Li et~al.(2017)Li, Deng, and Du]{li2017reliable}
Shan Li, Weihong Deng, and JunPing Du.
\newblock Reliable crowdsourcing and deep locality-preserving learning for expression recognition in the wild.
\newblock In \emph{Proceedings of the IEEE conference on computer vision and pattern recognition}, pages 2852--2861, 2017.

\bibitem[Li et~al.(2023)Li, Meinel, and Yang]{li2023generalized}
Ziyun Li, Christoph Meinel, and Haojin Yang.
\newblock Generalized categories discovery for long-tailed recognition.
\newblock \emph{arXiv preprint arXiv:2401.05352}, 2023.

\bibitem[Liu et~al.(2022)Liu, Dai, Feng, Wang, Yin, Zeng, and Shan]{liu2022mafw}
Yuanyuan Liu, Wei Dai, Chuanxu Feng, Wenbin Wang, Guanghao Yin, Jiabei Zeng, and Shiguang Shan.
\newblock Mafw: A large-scale, multi-modal, compound affective database for dynamic facial expression recognition in the wild.
\newblock In \emph{Proceedings of the 30th ACM International Conference on Multimedia}, pages 24--32, 2022.

\bibitem[Liu et~al.(2024)Liu, Huang, Liu, Zhan, Chen, and Chen]{liu2024open}
Yuanyuan Liu, Yuxuan Huang, Shuyang Liu, Yibing Zhan, Zijing Chen, and Zhe Chen.
\newblock Open-set video-based facial expression recognition with human expression-sensitive prompting.
\newblock \emph{arXiv preprint arXiv:2404.17100}, 2024.

\bibitem[Loshchilov and Hutter(2016)]{loshchilov2016sgdr}
Ilya Loshchilov and Frank Hutter.
\newblock Sgdr: Stochastic gradient descent with warm restarts.
\newblock \emph{arXiv preprint arXiv:1608.03983}, 2016.

\bibitem[Luo et~al.(2024)Luo, Du, Shi, Chen, Zhao, and Huang]{luo2024contextuality}
Tingzhang Luo, Mingxuan Du, Jiatao Shi, Xinxiang Chen, Bingchen Zhao, and Shaoguang Huang.
\newblock Contextuality helps representation learning for generalized category discovery.
\newblock \emph{arXiv preprint arXiv:2407.19752}, 2024.

\bibitem[MacQueen et~al.(1967)]{macqueen1967some}
James MacQueen et~al.
\newblock Some methods for classification and analysis of multivariate observations.
\newblock In \emph{Proceedings of the fifth Berkeley symposium on mathematical statistics and probability}, pages 281--297. Oakland, CA, USA, 1967.

\bibitem[Mollahosseini et~al.(2017)Mollahosseini, Hasani, and Mahoor]{mollahosseini2017affectnet}
Ali Mollahosseini, Behzad Hasani, and Mohammad~H Mahoor.
\newblock Affectnet: A database for facial expression, valence, and arousal computing in the wild.
\newblock \emph{IEEE Transactions on Affective Computing}, 10\penalty0 (1):\penalty0 18--31, 2017.

\bibitem[Pu et~al.(2023)Pu, Zhong, and Sebe]{pu2023dynamic}
Nan Pu, Zhun Zhong, and Nicu Sebe.
\newblock Dynamic conceptional contrastive learning for generalized category discovery.
\newblock In \emph{Proceedings of the IEEE/CVF Conference on Computer Vision and Pattern Recognition}, pages 7579--7588, 2023.

\bibitem[Rastegar et~al.(2024)Rastegar, Doughty, and Snoek]{rastegar2024learn}
Sarah Rastegar, Hazel Doughty, and Cees Snoek.
\newblock Learn to categorize or categorize to learn? self-coding for generalized category discovery.
\newblock \emph{Advances in Neural Information Processing Systems}, 36, 2024.

\bibitem[Ruan et~al.(2021)Ruan, Yan, Lai, Chai, Shen, and Wang]{ruan2021feature}
Delian Ruan, Yan Yan, Shenqi Lai, Zhenhua Chai, Chunhua Shen, and Hanzi Wang.
\newblock Feature decomposition and reconstruction learning for effective facial expression recognition.
\newblock In \emph{Proceedings of the IEEE/CVF conference on computer vision and pattern recognition}, pages 7660--7669, 2021.

\bibitem[Tan et~al.(2019)Tan, Liu, Ambrose, Tulig, and Belongie]{tan2019herbarium}
Kiat~Chuan Tan, Yulong Liu, Barbara Ambrose, Melissa Tulig, and Serge Belongie.
\newblock The herbarium challenge 2019 dataset.
\newblock \emph{arXiv preprint arXiv:1906.05372}, 2019.

\bibitem[Tian et~al.(2011)Tian, Kanade, and Cohn]{tian2011facial}
Yingli Tian, Takeo Kanade, and Jeffrey~F Cohn.
\newblock Facial expression recognition.
\newblock \emph{Handbook of face recognition}, pages 487--519, 2011.

\bibitem[Vaze et~al.(2021)Vaze, Han, Vedaldi, and Zisserman]{vaze2021open}
Sagar Vaze, Kai Han, Andrea Vedaldi, and Andrew Zisserman.
\newblock Open-set recognition: A good closed-set classifier is all you need?
\newblock 2021.

\bibitem[Vaze et~al.(2022)Vaze, Han, Vedaldi, and Zisserman]{vaze2022generalized}
Sagar Vaze, Kai Han, Andrea Vedaldi, and Andrew Zisserman.
\newblock Generalized category discovery.
\newblock In \emph{Proceedings of the IEEE/CVF Conference on Computer Vision and Pattern Recognition}, pages 7492--7501, 2022.

\bibitem[Wang et~al.(2024{\natexlab{a}})Wang, Vaze, and Han]{wang2024hilo}
Hongjun Wang, Sagar Vaze, and Kai Han.
\newblock Hilo: A learning framework for generalized category discovery robust to domain shifts.
\newblock \emph{arXiv preprint arXiv:2408.04591}, 2024{\natexlab{a}}.

\bibitem[Wang et~al.(2024{\natexlab{b}})Wang, Vaze, and Han]{wang2024sptnet}
Hongjun Wang, Sagar Vaze, and Kai Han.
\newblock Sptnet: An efficient alternative framework for generalized category discovery with spatial prompt tuning.
\newblock In \emph{International Conference on Learning Representations (ICLR)}, 2024{\natexlab{b}}.

\bibitem[Wang et~al.(2020)Wang, Peng, Yang, Lu, and Qiao]{wang2020suppressing}
Kai Wang, Xiaojiang Peng, Jianfei Yang, Shijian Lu, and Yu Qiao.
\newblock Suppressing uncertainties for large-scale facial expression recognition.
\newblock In \emph{Proceedings of the IEEE/CVF conference on computer vision and pattern recognition}, pages 6897--6906, 2020.

\bibitem[Wen et~al.(2023)Wen, Zhao, and Qi]{wen2023parametric}
Xin Wen, Bingchen Zhao, and Xiaojuan Qi.
\newblock Parametric classification for generalized category discovery: A baseline study.
\newblock In \emph{Proceedings of the IEEE/CVF International Conference on Computer Vision}, pages 16590--16600, 2023.

\bibitem[Wright(1990)]{wright1990speeding}
MB Wright.
\newblock Speeding up the hungarian algorithm.
\newblock \emph{Computers \& Operations Research}, 17\penalty0 (1):\penalty0 95--96, 1990.

\bibitem[Yang et~al.(2018)Yang, Ciftci, and Yin]{Yang_2018_CVPR}
Huiyuan Yang, Umur Ciftci, and Lijun Yin.
\newblock Facial expression recognition by de-expression residue learning.
\newblock In \emph{Proceedings of the IEEE Conference on Computer Vision and Pattern Recognition (CVPR)}, 2018.

\bibitem[Yu et~al.(2024)Yu, Wei, Cai, Zhao, Zhang, Wang, Xie, Zhu, Zhu, Liu, et~al.]{yu2024exploring}
Jun Yu, Zhihong Wei, Zhongpeng Cai, Gongpeng Zhao, Zerui Zhang, Yongqi Wang, Guochen Xie, Jichao Zhu, Wangyuan Zhu, Qingsong Liu, et~al.
\newblock Exploring facial expression recognition through semi-supervised pre-training and temporal modeling.
\newblock In \emph{Proceedings of the IEEE/CVF Conference on Computer Vision and Pattern Recognition}, pages 4880--4887, 2024.

\bibitem[Zhang et~al.(2023)Zhang, Khan, Shen, Naseer, Chen, and Khan]{zhang2023promptcal}
Sheng Zhang, Salman Khan, Zhiqiang Shen, Muzammal Naseer, Guangyi Chen, and Fahad~Shahbaz Khan.
\newblock Promptcal: Contrastive affinity learning via auxiliary prompts for generalized novel category discovery.
\newblock In \emph{Proceedings of the IEEE/CVF Conference on Computer Vision and Pattern Recognition}, pages 3479--3488, 2023.

\bibitem[Zhang et~al.(2024{\natexlab{a}})Zhang, Li, Liu, Deng, et~al.]{zhang2024leave}
Yuhang Zhang, Yaqi Li, Xuannan Liu, Weihong Deng, et~al.
\newblock Leave no stone unturned: Mine extra knowledge for imbalanced facial expression recognition.
\newblock \emph{Advances in Neural Information Processing Systems}, 36, 2024{\natexlab{a}}.

\bibitem[Zhang et~al.(2024{\natexlab{b}})Zhang, Yao, Liu, Qin, Wang, and Deng]{zhang2024open}
Yuhang Zhang, Yue Yao, Xuannan Liu, Lixiong Qin, Wenjing Wang, and Weihong Deng.
\newblock Open-set facial expression recognition.
\newblock \emph{arXiv preprint arXiv:2401.12507}, 2024{\natexlab{b}}.

\bibitem[Zhao and Han(2021)]{zhao2021novel}
Bingchen Zhao and Kai Han.
\newblock Novel visual category discovery with dual ranking statistics and mutual knowledge distillation.
\newblock \emph{Advances in Neural Information Processing Systems}, 34:\penalty0 22982--22994, 2021.

\bibitem[Zhao and Mac~Aodha(2023)]{zhao2023incremental}
Bingchen Zhao and Oisin Mac~Aodha.
\newblock Incremental generalized category discovery.
\newblock In \emph{Proceedings of the IEEE/CVF International Conference on Computer Vision}, pages 19137--19147, 2023.

\bibitem[Zhao et~al.(2023)Zhao, Wen, and Han]{zhao2023learning}
Bingchen Zhao, Xin Wen, and Kai Han.
\newblock Learning semi-supervised gaussian mixture models for generalized category discovery.
\newblock In \emph{Proceedings of the IEEE/CVF International Conference on Computer Vision}, pages 16623--16633, 2023.

\bibitem[Zhao et~al.(2016)Zhao, Gan, Wang, and Ji]{zhao2016facial}
Rui Zhao, Quan Gan, Shangfei Wang, and Qiang Ji.
\newblock Facial expression intensity estimation using ordinal information.
\newblock In \emph{Proceedings of the IEEE conference on computer vision and pattern recognition}, pages 3466--3474, 2016.

\end{thebibliography}
}

\maketitlesupplementary

\section{Experiments}
\subsection{Implementation Details}
We adopt DINO ViT-B/16 as the feature extractor, fine-tuning only the final Transformer blocks, which outputs 
\textit{\textbf{768}}-dimensional representations. The main head maps normalized bottleneck features (
\textit{\textbf{256}}-dimensional) to the classification space using a weight-normalized linear layer. The auxiliary head, designed for adversarial training, includes a gradient reversal layer followed by two fully connected layers with 2048 hidden units, GeLU activation, and dropout, enhancing robustness against biases.

\subsection{Dataset Details}\label{dataset}
We adopt the class division from~\cite{vaze2021open} utilizing 50\% of the images from these labeled classes as labeled instances in $D_L$. The remaining images from these classes are considered as the unlabeled data $D_U$. \textbf{RAF-DB}~\cite{li2017reliable} contains eleven compound expressions and seven basic expressions.

\begin{table}[h]  
\centering  
\caption{Specific category divisions.}  
\label{tab:datasets}  
\begin{tabular}{lcc}  
\toprule  
 &Old Classes &New Classes \\ 
\midrule  
RAF-DB-Basic  &$|\mathcal{Y}_l| = 4$  & $|\mathcal{Y}_u| - |\mathcal{Y}_l| = 3$  \\  
RAF-DB-Compound & $|\mathcal{Y}_l| = 7$ & $|\mathcal{Y}_u| - |\mathcal{Y}_l| = 11$ \\  
FerPlus       & $|\mathcal{Y}_l| = 4$ & $|\mathcal{Y}_u| - |\mathcal{Y}_l| = 4$  \\  
AffectNet     & $|\mathcal{Y}_l| = 4$ & $|\mathcal{Y}_u| - |\mathcal{Y}_l| = 4$  \\  
\bottomrule  
\end{tabular}  
\end{table}  

  
For GCD, we treat four out of the seven basic expressions as known categories, with three different selection methods. Notably, in the stage of discovering compound expressions, we use the basic expressions as known categories.  
The \textbf{RAF-DB-Basic} dataset is focused on basic expressions, while the \textbf{RAF-DB-Compound} dataset includes more complex expressions. We selected a subset of the basic expressions as old classes for the purpose of our experiments.  
\textbf{FERPlus}~\cite{barsoum2016training} is an extension of the FER2013 dataset~\cite{goodfellow2013challenges}. We selected a subset of the categories as old classes to discover the remaining new categories in this dataset.  
\textbf{AffectNet}~\cite{mollahosseini2017affectnet} is a large-scale facial expression recognition dataset. Similar to the selection strategy for FERPlus, we selected four categories as old classes in AffectNet to discover the remaining expressions.  


\textbf{The basis for partitioning.} We used the following partitioning means on the three datasets (RAF-DB-Basic, FerPlus, AffecNet): \textit{(Sur.+Fea.+Dis.+Hap.)} as known categories represent a more balanced partitioning. Relatively, \textit{(Sur.+Fea.+Dis.+Ang.)} represents more new categories and \textit{(Sur.+Hap.+Sad.+Neu.)} represents more old categories.
Please note that this is just one of the many approaches we have adopted for partitioning, and there are indeed numerous other ways to do so. In real-world deployments, partitioning can actually be done randomly.

\subsection{A More Detailed Ablation Study} 

In the main text, we have already conducted ablation experiments on the implicit and explicit components. In this section, we will perform a more detailed ablation of these components, as shown in Tab. \ref{ablation_v2}. When combining sample-level optimization with implicit debiasing, we achieve All ACC improvements of 9.0\% and 1.1\%, while combining distribution-level optimization with implicit debiasing results in All ACC improvements of 9.8\% and 1.3\%.

\begin{table}[t]
\centering
\caption{Ablation studies of different components of our method on the RAF-DB basic expression classes.}
\footnotesize
\label{ablation_v2}
\begin{tabular}{@{}ccccccccc@{}}
\toprule
\multicolumn{9}{c}{\textbf{Implicit Debiasing:} $\mathcal{L}_\text{ad}$, \textbf{Explicit Debiasing:} $\mathcal{L}_\text{bal}$, $\mathcal{L}_\text{cluster}$} \\
\toprule
$\mathcal{L}_\text{ad}$ & $\mathcal{L}_\text{bal}$ & $\mathcal{L}_\text{cluster}$ & \multicolumn{3}{c}{Sur.+Fea.+Dis.+Hap.} & \multicolumn{3}{c}{Sur.+Fea.+Dis.+Ang.} \\ \cmidrule(l){4-9} 
 &  &  & All & Old & New & All & Old & New \\ \midrule
 &  &  & 43.3 & 47.9 & 40.2  & 49.6 & 56.9 & 48.5\\
$\checkmark$ &  &  & 50.4 & 63.2 & 40.5 & 50.5 & 60.5 & 48.6\\
$\checkmark$ & $\checkmark$ &  & 52.3 & 65.3 & 40.4 & 50.7 & 63.4 & 47.5\\
$\checkmark$ & &   $\checkmark$& 53.1  & 71.4  & 40.6 & 50.9 & 64.1 & 48.7\\

 & $\checkmark$  & $\checkmark$ & 49.8 &61.1 & 40.7 & 51.5 & 63.2 & 50.3 \\
$\checkmark$ & $\checkmark$ & $\checkmark$ & 53.3 & 71.3 & 41.0 & 52.4 & 66.3 & 50.2 \\ 
\bottomrule
\end{tabular}
\end{table}

\begin{figure}[t]
    \centering
    \includegraphics[width=1\linewidth]{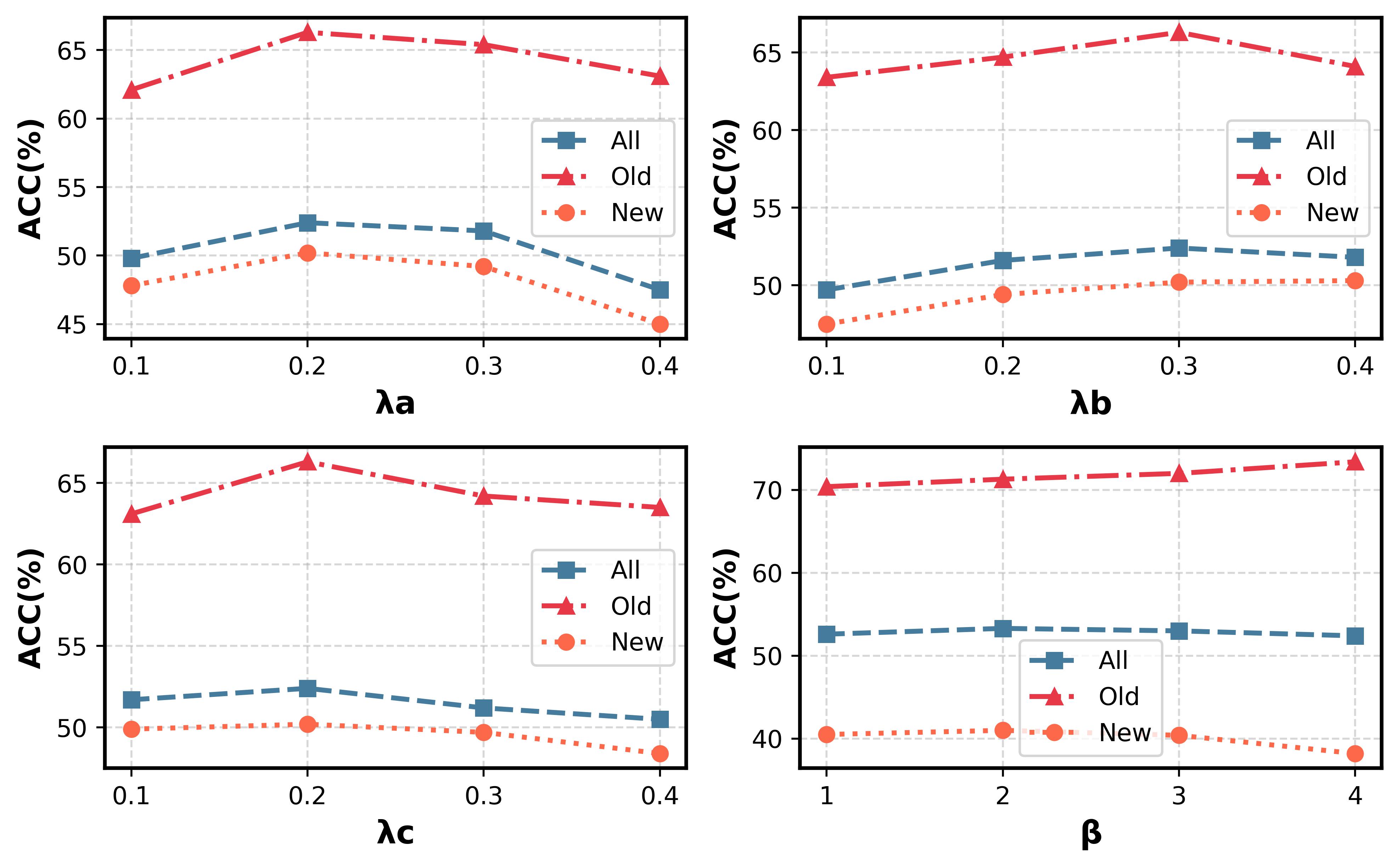}
    \caption{Hyperparametric analysis. $\lambda_a$, $\lambda_b$, and $\lambda_c$ represent the weights of the three losses, and $\beta$ represents the coefficients of the regular terms in the $\mathcal{L}_\text{cluster}$.}
    \label{Hyperparameters}
\end{figure}

\begin{table}[h]
\centering
\caption{Ablation study on the impact of using the dynamic adjustment term \( e_i \) in \( \mathcal{L}_{\text{bal}} \) on the RAF-DB dataset.}
\footnotesize
\begin{tabular}{@{}c|ccc|ccc@{}}
\toprule
 & \multicolumn{3}{c|}{Sur.+Fea.+Dis.+Hap.} & \multicolumn{3}{c}{Sur.+Fea.+Dis.+Ang.} \\ 
\cmidrule(l){2-7} 
\textbf{\( \mathcal{L}_{\text{bal}} \)} & \textbf{All} & \textbf{Old} & \textbf{New} & \textbf{All} & \textbf{Old} & \textbf{New} \\ 
\midrule
w/o \( e_i \) & 52.3 & 72.6  & 38.6 & 50.8 & 64.1 & 48.7 \\
w/ \( e_i \) & \textbf{53.3} & \textbf{71.3} & \textbf{41.0 }& \textbf{52.4} &\textbf{ 66.3} & \textbf{50.2} \\ 
\bottomrule
\end{tabular}
\label{ablation_ei}
\end{table}

\begin{figure*}[t]
    \centering
    \includegraphics[width=0.8\linewidth]{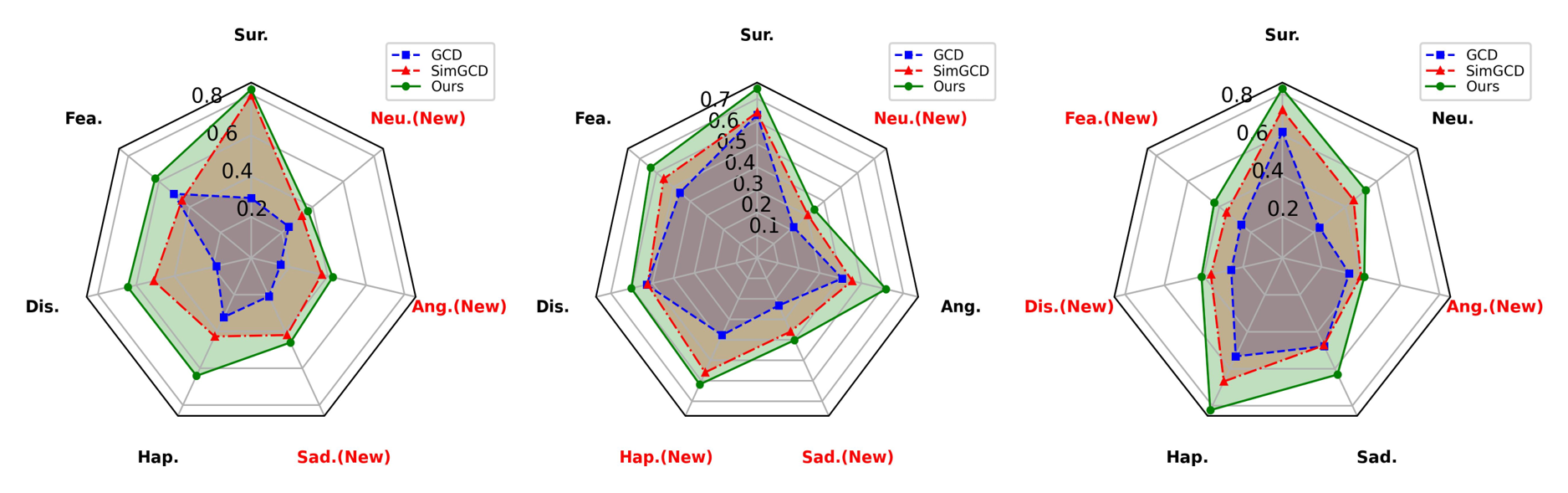}
    \caption{The ACC of single expression. Our approach effectively eliminates the bias caused by the introduction of new categories, and is able to better focus  minority expressions (\textit{e.g.}, Ang., Neu.).}
    \label{fig:enter-label}
\end{figure*}

\begin{figure*}[h]
    \centering
    \includegraphics[width=0.88\linewidth]{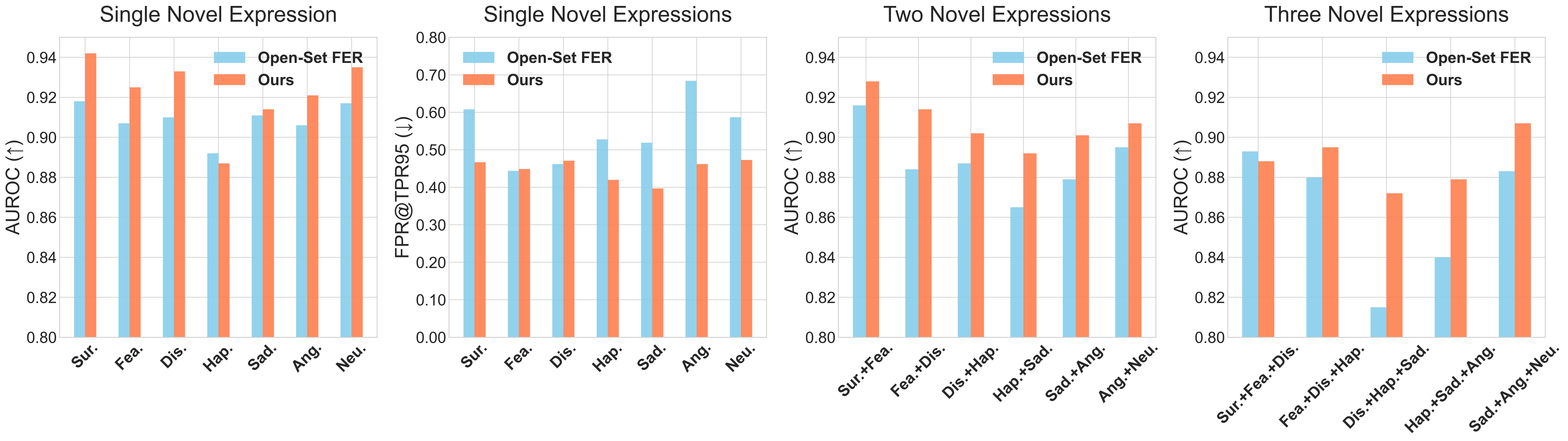}
    \caption{Comparison with  open-set FER~\cite{zhang2024open} in the open-set scenario. }
    \label{fig:open-set}
\end{figure*}
\begin{figure*}[h]
    \centering
    \includegraphics[width=0.8\linewidth]{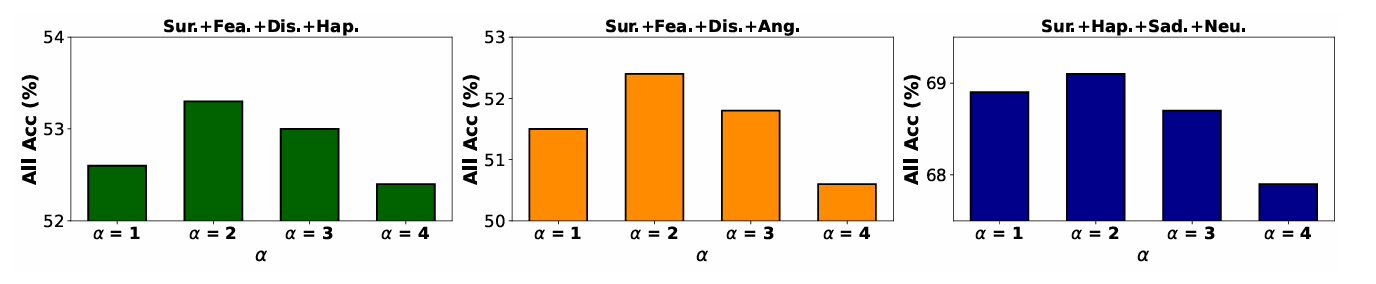}
    \caption{Category discovery accuracy (ACC) on RAF-DB. The effect of the hyperparameter $\alpha$.}
    \label{Hyperparameter_a}
\end{figure*}

\begin{table*}[h]
\caption{\textbf{Category discovery accuracy (ACC) on RAF-DB containing only known expressions. We test the FER method Ada-CM under GCD setting. $\triangle$ represents the improvement accuracy.}}
\centering
\footnotesize
\begin{tabular}{@{}lccccccccc@{}}
\toprule
& \multicolumn{3}{c}{Sur.+Fea.+Dis.+Hap.} & \multicolumn{3}{c}{Sur.+Fea.+Dis.+Ang.} & \multicolumn{3}{c}{Sur.+Hap.+Sad.+Neu.} \\ \cmidrule(l){2-4} \cmidrule(l){5-7} \cmidrule(l){8-10}
Method       & All     & Old     & New     & All     & Old     & New     & All     & Old     & New     \\ \midrule
Ours         & \textbf{53.3} &\underline{71.3}& \textbf{41.0} & \textbf{52.4} &  \textbf{66.3} & \textbf{50.2} & \textbf{69.1}&  \textbf{77.8} & \textbf{41.9}\\
\hline
Ada-CM~\cite{li2022towards}\footnotesize{\textcolor{gray}{[CVPR'22]}}         & 11.7 &18.2& 7.3 & 7.5 &  10.4 & 3.9 & 12.4&  9.7 & 14.8\\
Ada-CM+ our debias framework& 17.4 &29.2& 9.4 & 12.3 &  16.4 & 7.4 & 14.5&  11.4 & 18.6\\
$\triangle$      &+5.7 &+11.0& +2.1 & +4.8 &  +6.0 & +3.5 & +2.1&  +1.7 & +3.8\\
\bottomrule
\end{tabular}
\label{Ada-CM}
\end{table*}

\begin{figure}[h]
    \centering
    \includegraphics[width=0.7\linewidth]{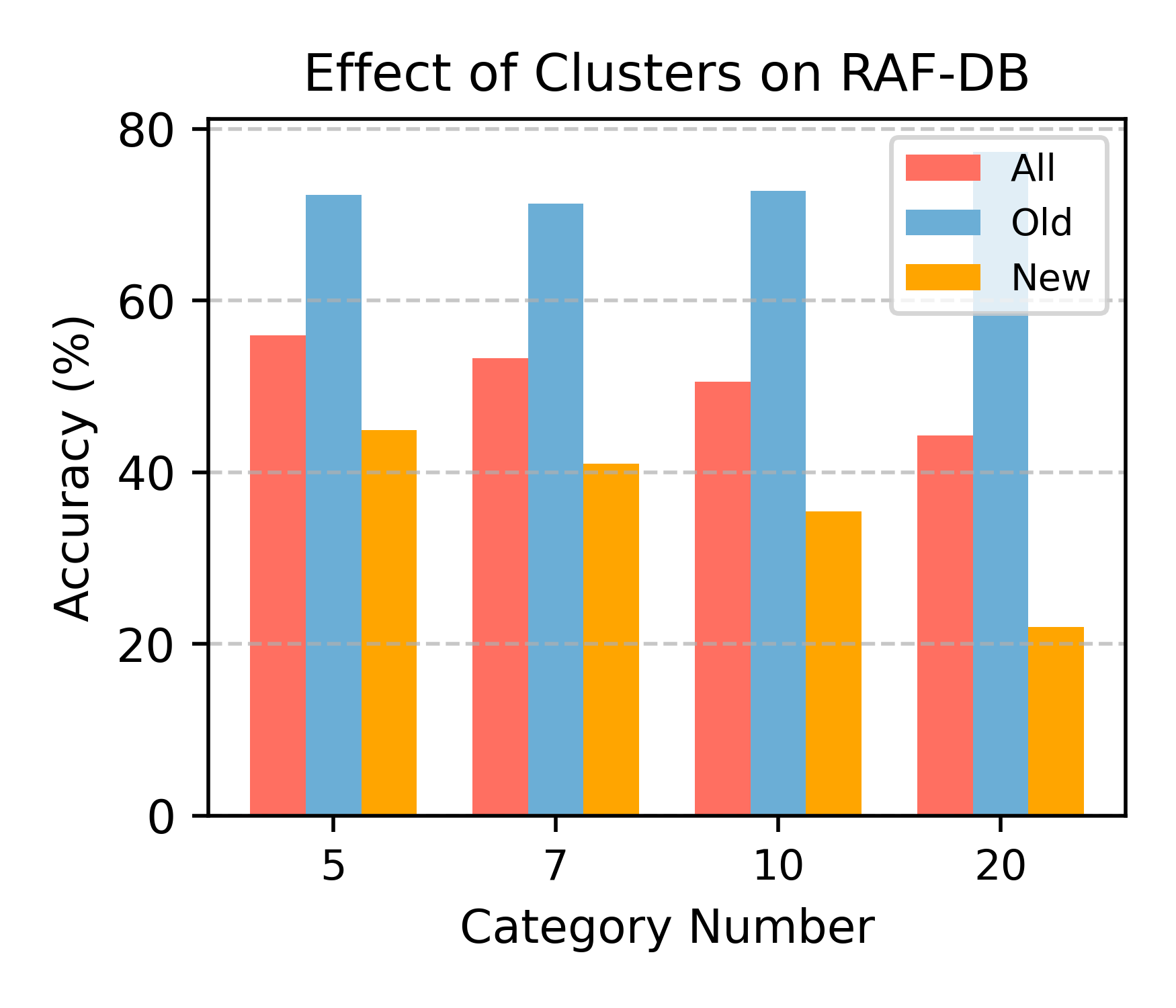}
    \caption{Effect of different number of clusters in our DIG-FACE on RAF-DB.}
    \label{category_number}
\end{figure}


\subsection{The Influence of the Hyperparameters}\label{analysis}
Our method has three main hyperparameters, which are the weight of Loss Functions. We show the results of our model in Fig. \ref{Hyperparameters} on RAF-DB  by tuning $\lambda_a$, $\lambda_b$ and $\lambda_c$ with different values. In general our model has good stability for different hyperparameters. We note that if too much weight is applied to the adversarial bias loss, the model performance decreases. Since we only use the adversarial loss to estimate the upper bound of bias, paying excessive attention to this loss can make the model difficult to optimize by other components. In addition $\mathcal{L}_{bal}$ should preferably not be weighted more than the contrastive learning loss $\mathcal{L}_{rep}$, which can lead to limitations in the model's ability to learn feature representations. Furthermore, $\mathcal{L}_{cluster}$ aims to further improve the expression-discriminability of the feature boundaries and we find that the weight that most improves model recognition is 0.2. Here, we have only roughly obtained an optimal combination of loss weights without conducting a precise analysis. Our main focus was on exploring the effectiveness of the components. The hyperparameters $\alpha$ and $\beta$ are analyzed as shown in Fig. \ref{Hyperparameter_a} and Fig. \ref{Hyperparameters}, taking values of 0.2 and 0.1, respectively. We also investigated the effect of $e_i$ in $\mathcal{L}_{bal}$, and Tab. \ref{ablation_ei} shows results with/without $e_i$, illustrating that adaptive changes in smoothness are important.

\subsection{Comparision with FER Method}  
To verify the generalizability of our method over previous FER approaches, we selected Ada-CAM~\cite{li2022towards}, a semi-supervised method adapted for the G-FACE task, as a baseline. Ada-CAM achieves an average \textbf{All ACC} of only 10.5\%. By integrating our DIG-FACE framework into Ada-CAM, we achieve notable improvements of 5.7\%, 4.8\%, and 2.1\% in \textbf{All ACC} across different settings, demonstrating the superior adaptability of our approach. Detailed results are provided in the Tab. \ref{Ada-CM}.
Additionally, adapting to the recent Open-set FER scenario introduced by \cite{zhang2024open} \textit{et al.}, we used AUROC (higher is better) and FPR@TPR95 (lower is better) as metrics. Fig. \ref{fig:open-set} shows the results, 
DIG-FACE achieved higher AUROC, indicating strong performance in detecting new categories, and lower FPR@TPR95, reflecting greater decision confidence.

\subsection{Sensitivity of the number of categories}
In this study, we use a parametric classifier approach assuming that the categories are known~\cite{wen2023parametric,wang2024sptnet}. In this section, we analyze the effect of different number of categories on ACC. Fig. \ref{category_number} shows the results, and since FER is a highly imbalanced scenario, increasing the number of output categories of the output header appears as a rise in the old categories. In future work, we will further explore methods that do not have a number of categories a priori.

\section{Pseudo Code} 
 We summarize the pipeline of \textbf{DIG-FACE} in Algorithm \ref{algorithm1}. \\
 \begin{algorithm}[h]
 \footnotesize
	\caption{The proposed DIG-FACE framework}
	\textbf{Input}: Train set $\mathcal{D} = \{\mathcal{D}_L \cup \mathcal{D}_U\}$, feature exactor $\psi$, main head $h$, auxiliary head $h_a$, train epoch $T$, warm-up epochs $w$.
	
	\textbf{Output}: Trained model parameter $\mathcal{S}$.
	
	\textbf{Initialize}: Load DINO ~\cite{caron2021emerging} pre-trained parameters for the backbone. 
	
	\begin{algorithmic}
		\FOR {$\text{epoch} = 0,\cdots,T-1$}
		
		\STATE Calculate the predictions of the main head with Eq. (13) and Eq. (9); 
		 
		\STATE Compute the supervised loss $\mathcal{L}_{\text{rep}}^{s}(\mathcal{D}_L)$ and self-supervised loss $\mathcal{L}_{\text{rep}}^{u}(\mathcal{D})$;
        \STATE Feed $\mathcal{D}$ to $h(\psi(\cdot))$ and $h_a(\text{GRL}(\psi(\cdot)))$ \COMMENT{GRL: Gradient Reversal Layer}  
        \STATE Calculate the Adversarial Loss $\mathcal{L}_{ad}$ with Eq. (8);  

		\IF {$epoch \geq w$} 
		\STATE Feed $\mathcal{D}_L$ to $\psi(\cdot)$ and then obtain features;
		\STATE Compute the Cluster Loss $L_{cluster}(\psi(\mathcal{D}_L)$;
		\ENDIF
		\ENDFOR
	\end{algorithmic}
	\label{algorithm1}
\end{algorithm}
After training stage, we would typically perform the GCD~\cite{vaze2022generalized}  metric evaluation directly in the program. The test strategy is summarized in Algorithm \ref{algorithm2}.

\begin{algorithm}[h]
	\caption{The test stage strategy of DIG-FACE}
	\textbf{Input}: The test set $\mathcal{D}^t = \left\{(\boldsymbol{x}_j, {y}_j)\right\}\in \mathcal{X}\times \mathcal{Y}$, trained DIG-FACE model $f_{s}$.
	
	\textbf{Output}: Classification accuracy $\mathrm{ACC}$.
	
	\begin{algorithmic}
		\FOR {$\boldsymbol{x}_i \in \mathcal{D}^t$} 
		\STATE Obtain the feature of $\boldsymbol{x}_j$ via $f_{s}(\boldsymbol{x}_j)$
		\ENDFOR
		\STATE Calculate the optimal assignment between clusters and categories by Hungarian algorithm ~\cite{wright1990speeding};
		\STATE Compute the test accuracy $\mathrm{ACC}$ based on the optimal assignment.

	\end{algorithmic}
	\label{algorithm2}
\end{algorithm}	
\onecolumn  
\clearpage
\section{Theory}

\subsection{Notation}

G-FACE learning scenario with a model $\mathcal{H}_{h,\psi}$ trained on both a labeled dataset $\mathcal{D}_L$ containing $\mathcal{N}$ categories ($\mathcal{N}$ represents the number of old categories) and an unlabeled dataset $\mathcal{D}_U$ containing $\mathcal{N+M}$ categories ($\mathcal{M}$ represents the number of new categories). In this task, characterized by both a large and finite amount of data, there exists a correct category space $\mathcal{F}$ for $\mathcal{D}_L$ and a potentially correct category space $\mathcal{\hat{F}}$ for $\mathcal{D}_U$. The model $\mathcal{H}_{h,\psi}$, comprising a feature exactor $\psi$ and a projection head $h$, represents the current operational model, while $\mathcal{H}_{h^*,\psi^*}$ denotes the optimal model. 
\subsection{Metric of Bias}
\noindent\textbf{Definition 1.} With a sufficiently large amount of data, we define $\xi(\cdot,\cdot)$ as a measure of the difference between the model's predictions and the ground truth. where the model's predictions go through a \textbf{softmax} layer. Since there is a sufficient amount of data, we use the following metric:

\begin{equation}
    \xi(\mathcal{H}_{h,\psi}(x),\mathcal{F}(x)) = \sqrt{\sum_{i=1}^{n} \left | \right | \mathcal{H}_{h,\psi}(x_i)-\mathcal{F}(x_i)\left | \right |^2} ,
\end{equation}
Next we need to prove that it satisfies the  metric properties.

\begin{Proof}\label{definition1}
To simplify the calculations:
\begin{equation}
    \mathcal{H}_{h,\psi}(x) = \mathcal{H}(x) \And \mathcal{H^*}_{h^*,\psi^*}(x) = \mathcal{H^*}(x),
\end{equation}
\begin{equation}
 \|\mathcal{H}(x)-\mathcal{F}(x)\|^2 =  \|\mathcal{H}(x)-\mathcal{H^*}(x)\|^2 + \|\mathcal{H^*}(x)-\mathcal{F}(x)\|^2 + 2(\mathcal{H}(x)-\mathcal{H^*}(x))\cdot(\mathcal{H^*}(x)-\mathcal{F}(x)),
\end{equation}

\begin{equation}
\begin{aligned}
    &\sum_{x=1}^{\left | D \right | } \left | \right | \mathcal{H}(x)-\mathcal{F}(x)\left | \right |^2 \\
    &=  \sum_{x=1}^{\left | D \right | }\|\mathcal{H}(x)-\mathcal{H^*}(x)\|^2 +2\sum_{x=1}^{\left | D \right | }(\mathcal{H}(x)-\mathcal{H^*}(x))\cdot(\mathcal{H^*}(x)-\mathcal{F}(x))+\sum_{x=1}^{\left | D \right | }\|\mathcal{H^*}(x)-\mathcal{F}(x)\|^2,
\end{aligned}
\end{equation}
Since the cross terms may be negative, we need to safely estimate the upper bound, according to Cauchy-Buniakowsky-Schwarz Inequality:
\begin{equation}
  \left[ \sum_{x=1}^{\left | D \right | } (\mathcal{H}(x)-\mathcal{H^*}(x)) \cdot (\mathcal{H^*}(x)-\mathcal{F}(x)) \right]^2 \leq \left[ \sum_{x=1}^{\left | D \right | } \left\| \mathcal{H}(x)-\mathcal{H^*}(x) \right\|^2 \right] \cdot \left[ \sum_{x=1}^{\left | D \right | } \left\| \mathcal{H^*}(x)-\mathcal{F}(x) \right\|^2 \right],
\end{equation}
\begin{equation}
 \left| \sum_{x=1}^{\left | D \right | } (\mathcal{H}(x)-\mathcal{H^*}(x)) \cdot (\mathcal{H^*}(x)-\mathcal{F}(x)) \right| \leq \sqrt{\sum_{x=1}^{\left | D \right | } \left\| \mathcal{H}(x)-\mathcal{H^*}(x) \right\|^2} \cdot \sqrt{\sum_{x=1}^{\left | D \right | } \left\| \mathcal{H^*}(x)-\mathcal{F}(x) \right\|^2},
\end{equation}
This allows us to perform a safe deflation to estimate the upper bound:

\begin{equation}
    \begin{aligned}
    &\sum_{x=1}^{\left | D \right | } \left\| \mathcal{H}(x)-\mathcal{F}(x) \right\|^2 \\
    &= \sum_{x=1}^{\left | D \right | }\|\mathcal{H}(x)-\mathcal{H^*}(x)\|^2 +2\sum_{x=1}^{\left | D \right | }(\mathcal{H}(x)-\mathcal{H^*}(x))\cdot(\mathcal{H^*}(x)-\mathcal{F}(x))  +\sum_{x=1}^{\left | D \right | }\|\mathcal{H^*}(x)-\mathcal{F}(x)\|^2 \\
    &\leq  \sum_{x=1}^{\left | D \right | }\|\mathcal{H}(x)-\mathcal{H^*}(x)\|^2 +2\sqrt{\sum_{x=1}^{\left | D \right | } \left\| \mathcal{H}(x)-\mathcal{H^*}(x) \right\|^2} \cdot \sqrt{\sum_{x=1}^{\left | D \right | } \left\| \mathcal{H^*}(x)-\mathcal{F}(x) \right\|^2}  +\sum_{x=1}^{\left | D \right | }\|\mathcal{H^*}(x)-\mathcal{F}(x)\|^2,
    \end{aligned}
\end{equation}
Finally we get the inequality:

\begin{equation}
    \begin{aligned}
    &\sqrt{\sum_{x=1}^{\left | D \right | } \left\| \mathcal{H}(x)-\mathcal{F}(x) \right\|^2} \\
    &\leq  \sqrt{\sum_{x=1}^{\left | D \right | }\|\mathcal{H}(x)-\mathcal{H^*}(x)\|^2 +2\sqrt{\sum_{x=1}^{\left | D \right | } \left\| \mathcal{H}(x)-\mathcal{H^*}(x) \right\|^2} \cdot \sqrt{\sum_{x=1}^{\left | D \right | } \left\| \mathcal{H^*}(x)-\mathcal{F}(x) \right\|^2}+\sum_{x=1}^{\left | D \right | }\|\mathcal{H^*}(x)-\mathcal{F}(x)\|^2 }\\
    &=\sqrt{\sum_{x=1}^{\left | D \right | }\|\mathcal{H}(x)-\mathcal{H^*}(x)\|^2} +\sqrt{\sum_{x=1}^{\left | D \right | }\|\mathcal{H^*}(x)-\mathcal{F}(x)\|^2 },
    \end{aligned}
\end{equation}

So we have:
\begin{equation}
     \xi(\mathcal{H}(x),\mathcal{F}(x)) \leq \xi(\mathcal{H}(x),\mathcal{H^*}(x)) + \xi(\mathcal{H^*}(x),\mathcal{F}(x)),
\end{equation}
And it's clear that:
\begin{equation}
     \xi(\mathcal{H}(x),\mathcal{F}(x)) \geq 0 \And \xi(\mathcal{H}(x),\mathcal{F}(x)) = \xi(\mathcal{F}(x),\mathcal{H}(x)), 
\end{equation}
\end{Proof}
\subsection{Bounding the Bias}
\noindent\textbf{Definition 3 (\textit{F-discrepancy})}
    For the hypothesis space $\mathcal{R}$, we define the upper bound on the discrepancy between the current model $\mathcal{H}$ and  an arbitrary model hypothesis  $\mathcal{H'}$  on both labeled and unlabeled data. The \textit{F-discrepancy} is:
\begin{equation}
 \mathop{\Delta}(\mathcal{D}_U,\mathcal{D}_L)  = \sup_{\mathcal{H},\mathcal{H'} \in \mathcal{R}} \left| \xi_{\mathcal{D}_U}(\mathcal{H}, \mathcal{H'}) - \alpha\cdot\xi_{\mathcal{D}_L}(\mathcal{H}, \mathcal{H'}) \right|,
\end{equation}
where $\alpha$ is a tuning parameter, and in fact it is possible to change the weight of the prediction bias for labeled data by changing $\alpha$. Next we need to use \textit{F-dicrepancy} in the proof of lemma 1.

\noindent\textbf{Lemma 1 (Bounding New Category Bias).}
Let $\mathcal{R}$ be a hypothesis space under the DIG-FACE training process. $\mathcal{H}$, $\mathcal{H'}$, and $\mathcal{H^*}$ represent the current model, any arbitrary model hypothesis, and the model that minimizes the joint error, respectively.
In order not to access the labels of $\mathcal{D}_U$, we need to constrain them with labels on the $\mathcal{D}_L$ and theoretically define the \textbf{upper bound of bias} for new categories:
\begin{equation}
\xi^{new}_{\mathcal{D}_U}(\mathcal{H},\mathcal{\hat{F}}) \leq \frac{1}{1-\theta} \left(  (\alpha-\theta)\cdot\xi_{\mathcal{D}_L}(\mathcal{H},\mathcal{F}) + \mathop{\Delta}(\mathcal{D}_U,\mathcal{D}_L)+\lambda \right),
\end{equation}
\textit{Where} $\lambda = \alpha\cdot\xi_{\mathcal{D}_L}(\mathcal{H^*},\mathcal{F})+\xi_{\mathcal{D}_U}(\mathcal{H^*},\mathcal{\hat{F}})$. 

\begin{Proof}\label{Lemma 1}
    According to Definition 1, we have:
\begin{equation}
\begin{aligned}
&\mathbb{E}_{x \in D_U} \left[ \xi\left(\mathcal{H}(x), \mathcal{\hat{F}}(x)\right) \right] \\
&\quad= \mathbb{E}_{x \in D_U} \left[ \xi\left(\mathcal{H}(x), \mathcal{\hat{F}}(x)\right) \right] 
\leq \mathbb{E}_{x \in D_U} \left[ \xi\left(\mathcal{H^*}(x), \mathcal{\hat{F}}(x)\right) \right] 
+\mathbb{E}_{x \in D_U} \left[ \xi\left(\mathcal{H}(x), \mathcal{H^*}(x)\right) \right],
\end{aligned}
\end{equation}
So we have:
\begin{equation}
    \xi_{D_U}(\mathcal{H},\mathcal{\hat{F}})\leq\xi_{D_U}(\mathcal{H},\mathcal{H^*}) +\xi_{D_U}(\mathcal{H^*},\mathcal{\hat{F}}),
\end{equation}
$D_U$ contains both new and old categories:
\begin{equation}
    \begin{aligned}
        &\xi_{D_U}(\mathcal{H},\mathcal{\hat{F}}) = (1-\theta)\xi^{new}_{D_U}(\mathcal{H},\mathcal{\hat{F}}) + \theta\xi^{old}_{D_U}(\mathcal{H},\mathcal{\hat{F}}),
    \end{aligned}
\end{equation}
According to Definition 2, we have:
\begin{equation}\label{Eq31}
   \xi_{\mathcal{D}_L}(\mathcal{H},\mathcal{F}) \leq  \xi^{old}_{\mathcal{D}_U}(\mathcal{H},\hat{\mathcal{F}}) \leq \xi_{\mathcal{D}_U}(\mathcal{H},\hat{\mathcal{F}}),
\end{equation}


Considering that we want to constrain the new categories of bias:
\begin{equation}
\begin{aligned}
        \xi_{D_U}(\mathcal{H},\hat{\mathcal{F}}) = (1-\theta)\xi^{new}_{D_U}(\mathcal{H},\hat{\mathcal{F}}) + \theta\xi^{old}_{D_U}(\mathcal{H},\hat{\mathcal{F}})\leq\xi_{D_U}(\mathcal{H},\mathcal{H^*}) +\xi_{D_U}(\mathcal{H^*},\hat{\mathcal{F}}),
\end{aligned}
\end{equation}
Adding and subtracting terms:
\begin{equation}
    \begin{aligned}
        (1-\theta)\xi^{new}_{D_U}(\mathcal{H},\hat{\mathcal{F}}) + \theta\xi^{old}_{D_U}(\mathcal{H},\hat{\mathcal{F}})\leq\xi_{D_U}(\mathcal{H},\mathcal{H^*}) +\xi_{D_U}(\mathcal{H^*},\hat{\mathcal{F}}) - \alpha\cdot\xi_{D_L}(\mathcal{H},\mathcal{H^*}) + \alpha\cdot\xi_{D_L}(\mathcal{H},\mathcal{H^*}),
    \end{aligned}
\end{equation}

Rearrange the equation:

\begin{equation}  
    \begin{aligned}  
        (1-\theta)\xi^{new}_{D_U}(\mathcal{H},\hat{\mathcal{F}})\leq  \xi_{D_U}(\mathcal{H^*},\hat{\mathcal{F}}) - \alpha\cdot \xi_{D_L}(\mathcal{H},\mathcal{H^*}) +  \xi_{D_U}(\mathcal{H},\mathcal{H^*})+ \alpha\cdot\xi_{D_L}(\mathcal{H},\mathcal{H^*})- \theta\xi_{\mathcal{D}_L}(\mathcal{H},\mathcal{F}),
    \end{aligned}  
\end{equation}

Based on $\xi(\cdot,\cdot)$ metric properties, further scaling is applied:
\begin{equation}
    \begin{aligned}
          &(1-\theta)\xi^{new}_{D_U}(\mathcal{H},\hat{\mathcal{F}})\\
          &\leq \xi_{D_U}(\mathcal{H},\mathcal{H^*}) - \alpha\cdot\xi_{D_L}(\mathcal{H},\mathcal{H^*}) +  \alpha\cdot\xi_{D_L}(\mathcal{H^*},\mathcal{F}) +\alpha\cdot\xi_{D_L}(\mathcal{H},\mathcal{F})-\theta\cdot  \xi_{D_L}(\mathcal{H},\mathcal{F})+\xi_{D_U}(\mathcal{H^*},\hat{\mathcal{F}}) \\  
        &\leq \left | \xi_{D_U}(\mathcal{H},\mathcal{H^*}) - \alpha\cdot\xi_{D_L}(\mathcal{H},\mathcal{H^*}) \right | +(\alpha-\theta)\cdot\xi_{D_L}(\mathcal{H},\mathcal{F}) + \lambda\\  
        &\leq \underbrace{\sup_{\mathcal{H},\mathcal{H'} \in \mathcal{R}}\left | \xi_{D_U}(\mathcal{H},\mathcal{H'}) - \alpha\cdot\xi_{D_L}(\mathcal{H},\mathcal{H'}) \right |}_{\mathop{\Delta}(\mathcal{D}_U,\mathcal{D}_L)} +(\alpha-\theta)\cdot \xi_{D_L}(\mathcal{H},\mathcal{F}) + \lambda,
    \end{aligned}
\end{equation}

where the constant $\lambda = \alpha\cdot\xi_{D_L}(\mathcal{H^*},\mathcal{F})+\xi_{D_U}(\mathcal{H^*},\hat{\mathcal{F}})$.

\end{Proof}

\twocolumn

\end{document}